\documentclass[manuscript, screen]{jair}

\JAIRAE{}
\JAIRTrack{Integration of Logical Constraints in Deep Learning}
\usepackage{bbm}
\usepackage{lipsum}
\newcommand\blfootnote[1]{%
  \begingroup
  \renewcommand\thefootnote{}\footnote{#1}%
  \addtocounter{footnote}{-1}%
  \endgroup
}
\usepackage{xcolor}
\definecolor{phtalogreen}{HTML}{123524}

\newcommand{\stt}[1]{{\small\texttt{#1}}}

\usepackage{subcaption}
\usepackage{xspace}
\newcommand{\RQinsighttraining}{\textbf{RQ1}\xspace}
\newcommand{\RQfeatures}{\textbf{RQ3}\xspace}
\newcommand{\RQinsightmultiagent}{\textbf{RQ2}\xspace}
\newcommand{\RQknowledge}{\textbf{RQ4}\xspace}

\newtheorem{assumption}{Assumption}
\newtheorem{definition}{Definition}

\begin{document}

%%
%% The "title" command has an optional parameter,
%% allowing the author to define a "short title" to be used in page headers.
%\title{JAIR Example Template}
\title[ILP for XRL]{Explaining Reinforcement Learning Agents via Inductive Logic Programming}

%%
%% The "author" command and its associated commands are used to define
%% the authors and their affiliations.
%% Of note is the shared affiliation of the first two authors, and the
%% "authornote" and "authornotemark" commands
%% used to denote shared contribution to the research and/or corresponding author.

\author{Celeste Veronese$^*{}^1$}

\email{celeste.veronese@univr.it}
\orcid{0009-0007-7461-4039}

\affiliation{%
  \institution{University of Verona}
  \city{Verona}
  \country{Italy}
}

\author{Edoardo Zorzi$^*$}
\email{edoardo.zorzi@univr.it}
\orcid{0000-0001-7090-9592}

\affiliation{%
  \institution{Sapienza University of Rome}
  \city{Rome}
  \country{Italy}
}
\affiliation{%
  \institution{University of Verona}
  \city{Verona}
  \country{Italy}
}

\author{Daniele Meli}
\email{daniele.meli@univr.it}
\orcid{0000-0002-3162-388X}

\affiliation{%
  \institution{University of Verona}
  \city{Verona}
  \country{Italy}
}

\author{Alessandro Farinelli}
\email{alessandro.farinelli@univr.it}
\orcid{0000-0002-2592-5814}

\affiliation{%
  \institution{University of Verona}
  \city{Verona}
  \country{Italy}
}

\renewcommand{\shortauthors}{Veronese et al.}

%%
%% The abstract is a short summary of the work to be presented in the
%% article.
\begin{abstract}\blfootnote{$^*$Equal Contribution}\footnotetext[1]{Corresponding author}
{\bf Background:}
    Explainable Reinforcement Learning (XRL) seeks to make Reinforcement Learning (RL) policies more transparent and interpretable, a key requirement in safety-critical and human-centric scenarios. However, it is mostly based on user studies, thus targeting the needs of a specific audience and lacking shared evaluation metrics.
    On the other hand, logic-based approaches within eXplainable Artificial Intelligence (XAI) provide compact, human-readable abstractions of decision-making. However, the systematic quantification of the explainability degree of logical representations remains an open problem.
    
{\bf Objectives:}
    This work aims to advance the state of the art in XRL by introducing objective and planning-oriented metrics for policy explainability in single-and multi-agent RL settings. At the same time, it contributes to the field of logics for XAI by providing a principled way to quantify the explainability of logical rules, moving beyond common-sense assessments and simple propositional fragments.
    
{\bf Methods:}
    We employ Inductive Logic Programming (ILP) to extract symbolic representations of RL policies and define a novel set of explainability metrics, including \textit{activation rate}, \textit{feature coverage}, \textit{syntactic distance} and \textit{semantic distance}. These metrics quantify alignment between symbolic rules and agent behavior, the role of features in decision-making, and the evolution of policies during training and across agents in single and Multi-Agent RL (MARL).
    
{\bf Results:}
    Experiments across different RL domains show that the proposed metrics highlight action-specific learning dynamics beyond global return, provide fine-grained insights into domain features beyond classical approaches for global feature importance estimation, and uncover coordination, specialization, and adaptation patterns in MARL. Moreover, they provide crucial insights for the transfer and generalization of action-specific policies.
    
{\bf Conclusions:}
    Our framework advances XRL by offering rigorous, objective, and interpretable metrics to evaluate symbolic policy representations. This contributes to understanding, debugging, and refining RL agents, paving the way for more robust and trustworthy applications in dynamic, safety-critical, and multi-agent environments.
\end{abstract}
%\begin{abstract}
%      A clear and well-documented \LaTeX\ document is presented as an
%  article formatted for publication by ACM in a conference proceedings
%  or journal publication. Based on the ``acmart'' document class, this
%  article presents and explains many of the common variations, as well
%  as many of the formatting elements an author may use in the
%  preparation of the documentation of their work.
%\end{abstract}

%% JAIR Note: 
%% Do not include ACM CCS Concepts or Keywords

%% To be updated by authors.
% \received{20 February 2007}
% \received[revised]{12 March 2009}
% \received[accepted]{5 June 2009}

%%
%% This command processes the author and affiliation and title
%% information and builds the first part of the formatted document.
\maketitle

\section{Introduction}\label{introduction}
Reinforcement Learning (RL) \cite{SuttonBarto2018} has emerged as a powerful paradigm for sequential decision-making problems, achieving remarkable success in complex domains such as robotics and autonomous systems \cite{kendall2019learning,levine2016end,silver2016mastering}. Despite these advances, a critical limitation persists: the lack of transparency and interpretability of the policies learned by RL agents. In many real-world scenarios, particularly those involving safety-critical applications or human interaction, understanding the rationale behind an agent's behavior is essential for fostering trust, ensuring compliance with domain constraints, and enabling debugging and validation processes.

Explainable Artificial Intelligence (XAI) seeks to address these challenges by developing methods that make Machine Learning (ML) models more interpretable to humans. Significant progress has been made in Explainable ML (XML) in recent years, especially in addressing the explainability of supervised learning models \cite{burkart2021survey}. Regardless of whether explainability is achieved through attributions (i.e. feature importance), counterfactuals, or logical rules, XAI mainly focuses on user-centric evaluations \cite{marques-silva2025logic} or quantitative metrics, such as the size of the explanation and its fidelity and robustness with respect to the original model, whose applicability is restricted to tabular-based ML \cite{perotti2024metrics}.

This topic is even less explored in the RL domain, where agents interact with complex, non-stationary environments and learn policies through trial and error over time \cite{milani2024survey}. 
Most existing approaches in this direction rely on user-centered criteria such as fidelity, comprehensibility, and preferability \cite{milani2024survey}, which play a crucial role in assessing the quality of explanations from the perspective of human understanding and user satisfaction. However, to move toward standardized and rigorous evaluation, it is important to complement these subjective dimensions with objective measures that can consistently assess explanations across different domains and user groups, ensuring both efficiency and effectiveness in their evaluation \cite{ras2022explainable, retzlaff2024human}.

The concept of \emph{feature importance}, which is typical of explainable ML, can be partially adapted to the RL context to identify the most relevant environmental features determining the agent’s actions \cite{sequeira2020interestigness}.
However, there is limited emphasis in Explainable RL (XRL) literature on an in-depth and objective (hence user-independent) analysis of the agent's training process \cite{ras2022explainable,retzlaff2024human}.
While recent frameworks \cite{amparore2021trust,nauta2023anecdotal} represent important steps towards defining more systematic explainability evaluation, they are primarily tailored to feature-attribution explanations in static supervised learning settings. 
Consequently, they are not directly applicable to policies derived from RL agents, particularly in dynamic environments. 
For instance, it is rarely explored how to evaluate the quality of the learned policy across different actions, which could be useful to understand on which decisions the agent is more confident or robust, an insight that could inform generalization potential and the need for policy refinement. Moreover, assessing the convergence of learned policies in multi-agent settings is crucial for understanding inter-agent relationships. \cite{foerster2016learning}. Another relevant aspect is analyzing changes in the agent’s policy under equal returns, which may reveal alternative, equally rewarding strategies that differ significantly from a semantical point of view. Finally, identifying the reasons behind non-stationarity in Multi-Agent Reinforcement Learning (MARL) can provide valuable insights into the dynamics and stability of such systems \cite{foerster2016learning}.

To address the aforementioned limitations, our work introduces a novel set of explainability metrics specifically designed for symbolic representations extracted from RL policies. We focus on the logical formalism since it represents a powerful and versatile tool to model the decision-making process of agents; as confirmed by the growing interest of the research community in neuro-symbolic (NeSy) integration that combines neural network-based ML with knowledge representation and reasoning from symbolic approaches \cite{garcez2023neurosymbolic}. 
In this context, considerable effort has been devoted to extracting symbolic representations of policies, such as rule-based models and logical formulas, which offer a compact, human-readable abstraction of the agent's decision-making process \cite{darwiche2022quantifying,marques-silva2025logic}.
Several methods have been proposed to infer such symbolic representations from RL agents by observing their behavior \cite{meli2024learning,veronese2023inductive,furelos2021induction}. 
However, to the best of our knowledge, a critical gap still exists in the current literature: the lack of systematic and user-agnostic criteria to measure the level of explainability of the RL training process and the generated policy \cite{ras2022explainable,retzlaff2024human}.
 
In this context, we focus on evaluating logical policy representations expressed in the Answer Set Programming (ASP) formalism \cite{gelfond1988stable}, which is an established symbolic representation paradigm for autonomous agents \cite{meli2023logic}. 
We learn the ASP policy representations from RL policy evaluations via Inductive Logic Programming (ILP) \cite{law2015learning}, starting from a set of user-defined predicates describing the actions of the domain and the main environmental features.
Our proposed metrics go beyond mere feature importance and subjective criteria for explainability assessment by focusing on more advanced and planning-oriented crucial aspects of an RL policy, e.g., its robustness to the environmental conditions and generalization capabilities. 
We empirically show how these metrics can be effectively used to evaluate and explain the dynamic training process of the RL agents, and assess the robustness, reliability, and generalization capabilities of learned skills, both in single and multi-agent RL settings. 
In detail, we make the following contributions:
\begin{itemize}
    \item we define a novel set of objective, user-independent explainability metrics to quantify different aspects of RL agents. These metrics include the \emph{activation rate} of the learned ASP policy representation, that is, how much the symbolic rules correctly aligns to the actual behavior of the agent (useful to highlight which action-specific policies the agent is more confident on); \emph{feature coverage}, to quantify the impact of symbolic features in the decision-making process of the agent (similar to feature importance from XML); \emph{syntactic} and \emph{semantic distance} to measure how two symbolic (logical) policy representation differ between each other, thus highlighting potential evolution of the skills of the agent during training or convergence between multiple agents;
    \item we perform a thorough empirical study to evaluate the advantages that our metrics bring to XRL. Specifically, we consider different RL settings of different complexity and features, from single-agent RL (Intersection), to cooperative (RWARE) and contrastive (Simple Adversary) multi-agent RL, with inherent non-stationarity and dynamic interconnections between the learning processes of different agents;
    \item we analyze the impact of explainability criteria and metrics in light of their utility for optimal sequential decision-making. Specifically, we show that our metrics can be employed to assess the generalization capabilities of the sub-policies of the agent for specific actions (via the activation rate), and to identify the learning of specific skills or the emergence of multi-agent agreement (via the semantic and syntactic distances).
\end{itemize}
The remainder of the paper is organized as follows: in Section \ref{sec:related_works}, we review recent research in XAI, focusing on XRL and, in particular, on the use of logics and formal methods to enhance explainability. Then, in Section \ref{sec:background} we introduce relevant background and notation about Markov Decision Processes (MDPs), ASP, and ILP. In Section \ref{sec:methodology} we present the proposed explainability metrics, together with the process of learning logical policy representations to be evaluated. We then perform our extensive empirical evaluation in Section \ref{sec:experiments}. Finally, in Section \ref{sec:conclusions} we summarize the obtained results and discuss the advantages and limitations of our work, also analyzing possible future extensions.
\section{Related Works}\label{sec:related_works}
%XAI is an interdisciplinary research field that aims to make AI systems more transparent, interpretable, and trustworthy to human users \cite{dwivedi2023explainable}. In particular, explainability is crucial in decision-making and safety-critical systems, where understanding the rationale behind the behaviour of the agent is essential for human oversight or collaboration.
%In this regard, logic-based methods have also emerged as a valuable tool to support formal, structured, and verifiable explanations, enabling the representation of agent behavior, decision processes, and learned knowledge in interpretable symbolic forms \cite{darwiche2022quantifying,marques-silva2025logic}.
%Within this broad landscape, RL poses unique challenges due to the dynamic and often completely black-box nature of learned policies. The field of XRL has thus emerged, focusing on enhancing the interpretability of RL agents and their decision-making processes. 
%Nonetheless, existing approaches to explainable decision-making and XRL either provide interpretable policy representations only for specific user groups, lacking an objective and measurable assessment of the explanations, or investigate only limited aspects of the decision-making process, e.g., feature importance for the policy output \cite{retzlaff2024human}.

Explainable Artificial Intelligence (XAI) aims to improve the transparency and interpretability of AI systems \cite{dwivedi2023explainable}. Within this landscape, logic-based approaches have gained increasing attention as a means to provide structured and formally grounded explanations of learned models. In particular, symbolic representations allow the behavior and decision processes of AI agents to be expressed in interpretable and verifiable forms \cite{darwiche2022quantifying,marques-silva2025logic}.
Applying explainability techniques to Reinforcement Learning (RL) remains challenging due to the dynamic nature of the environments and the black-box characteristics of many learned policies. As a result, the field of Explainable Reinforcement Learning (XRL) has developed a variety of approaches aimed at improving the interpretability of RL agents and their decision-making processes. However, many existing methods either focus on interpretable policy representations tailored to specific user groups or analyze limited aspects of the learned policy, such as feature importance for the policy output \cite{retzlaff2024human}, without providing objective and systematic criteria for evaluating explanations.

In the following, we analyze the related works in three main areas: (i) explainability in decision-making, focusing on post-hoc and behavior summarization methods; (ii) explainability in RL, addressing approaches aimed at global policy-level understanding; and (iii) logic-based methods for explainability, which leverage a formal symbolic representation of the domain to enhance the transparency and verifiability of AI models.
Our work lies at the intersection of the second and last research areas, focusing specifically on logic-based explanations derived from execution traces of RL agents and introducing evaluation metrics for the resulting symbolic policy approximations.

\subsection{Explainability in Decision-Making Systems}
Explainability in decision-making often aims to provide users with understandable justifications of the agent's choices or strategies. 
Several approaches focus on post-hoc explainability methods that analyze and explain agents' decisions without modifying the underlying policy, thus being applied to any kind of decision-making agent. 
For example, saliency-based techniques \cite{greydanus2018visualizing} generate visual explanations by highlighting important regions in the agent's observation space, while methods such as \cite{zahavy2016graying} employ t-distributed Stochastic Neighbor Embedding (t-SNE) to map high-dimensional state representations into interpretable two-dimensional spaces. However, these approaches often suffer from computational overhead and lack accessibility for non-expert users.
Efforts to summarize agent behavior to highlight key moments or states during training have also gained attention. For example, \cite{amir2018summarizing} proposes strategies to identify critical states for policy summarization, while \cite{huang2018establishing} emphasizes critical decision points that help users develop appropriate trust in the agent. Extending these ideas, \cite{sequeira2020interestigness} introduces a domain-agnostic framework that uses different \emph{interestingness} criteria to generate summaries of agent interactions during training, leveraging interaction data, such as state visit counts, action frequencies, transition probabilities, and value estimates. These approaches, however, heavily rely on user studies as their only evaluation method, introducing challenges related to the scalability and consistency of the produced explanations.

In response to this limitation, recent evaluation frameworks have proposed more principled approaches for assessing XAI methods. The Co-12 framework \cite{nauta2023anecdotal} identifies twelve core properties that explanations should satisfy, such as correctness, completeness, and consistency, and systematically categorizes evaluation methodologies for each. Similarly, the LEAF framework \cite{amparore2021trust} proposes a set of metrics for quantitatively evaluating local explanation faithfulness and stability, including local fidelity, reiteration similarity, and prescriptivity. While these efforts represent significant advances toward standardized evaluation, they predominantly focus on feature-attribution methods and ignore other relevant aspects of policy evaluation, e.g., the diversity of skills and curricula, policy robustness and generalization, and multi-agent coordination and agreement.

\subsection{Explainability in Reinforcement Learning}
Recently, explainability in decision-making has been extended to RL, with the goal of enhancing the transparency of the agent's learning process. 
To this aim, for example, causal reasoning has been integrated into RL agents' environment models \cite{madumal2020explainable}, in order to generate counterfactual explanations. Similarly, reward decomposition approaches \cite{anderson2019explaining,juozapaitis2019explainable} clarify trade-offs between different reward components, while hierarchical RL structures have been leveraged to promote modular, interpretable sub-policies \cite{rietz2022hierarchical}.
A significant body of work in XRL focuses on generating more interpretable representations of agents' policies or behaviors. For example, VIPER \cite{bastani2018viper} extracts decision tree policies from high-performing DRL agents, facilitating formal verification and human interpretability.  
Iterative Bounding Markov Decision Processes (IBMDPs) \cite{topin2021iterative} directly derive interpretable decision trees during the training of the agents. 
Similar approaches have been recently extended to explainability for MARL, though this is still an emergent and underexplored area of research \cite{boggess2025explanations}.
Crucially, the above methods mainly address explanations for specific user groups, lacking an objective assessment of the RL process and missing an in-depth analysis of the policy robustness, generalizability and diversity.

In addition to MDP-level explanations, research has focused on explaining the agents' policies at a global level, though this remains a less mature area \cite{milani2024survey}. For instance, trajectory-based methods \cite{amir2018summarizing} select representative trajectories based on state importance and diversity, while others identify key events \cite{dinu2022xai} or focus on highlighting the agent's interventions, that is, the moments where its actions critically influence the environment or task outcome \cite{jacq2022lazy}. Despite providing valuable insights, all of the aforementioned methods do not provide a quantitative assessment of the policy properties, and face challenges in non-stationary (e.g., multi-agent) domains, where notable events may be widely dispersed.

Model-based approaches also aim to enhance global policy interpretability. Clustering-based methods \cite{sreedharan2020tldr,mccalmon2022caps}, for example, provide high-level summaries of agent policies through abstract states or landmarks, facilitating human understanding of global agent strategies. However, being often derived from limited sets of trajectories or features, such methods typically lack formal mechanisms to evaluate how closely the explanations reflect the true decision-making process or how well the explanations apply to unseen or novel situations. As a result, these methods often face difficulties in assessing the fidelity and generalization capabilities of the obtained explanations, which limits their reliability and usefulness.

Pursuing the same objective, \cite{koul2018LearningFS} converts Relational Neural Network policies into interpretable finite state machines using quantized bottleneck networks, and \cite{hasanbeig2021deepsynth} introduces methods to synthesize deterministic automata representing agent policies, ensuring both interpretability and formal guarantees without relying on recurrent architectures.
These works exemplify the growing interest in leveraging formal symbolic representations to enhance the transparency of learned policies. Building on this perspective, logic-based methods have emerged as a distinct research direction.

\subsection{Logics and Formal Methods for Explainability}

As testified by \cite{darwiche2022quantifying,koul2018LearningFS,hasanbeig2021deepsynth,caporese2026RLmeetsLP}, recent research trends frequently leverage logical and formal methods to enhance explainability in both planning and RL settings. By representing agent behaviors or learned policies using formal models such as automata or expressive logical formalisms, these approaches aim to generate precise, verifiable, and structured explanations. 
For instance, program synthesis approaches like PIRL \cite{verma2018programmatically} leverage symbolic representations to expressively represent the policy structure.
Within this line of research, recent works have also explored the application of computation tree logic (CTL) to enhance the explainability of online sequential planners such as Monte Carlo Tree Search (MCTS). For example, \cite{an2024enabling} proposes a CTL-based framework to explain MCTS-based planning decisions in the context of transportation services. Their approach allows non-technical users to pose factual, contrastive, and exploratory queries about the planner's actions, which are automatically translated into CTL formulas. The framework then verifies these formulas over the MCTS search tree and translates the results back into human-readable natural language explanations using predefined templates. 
Nevertheless, these approaches focus on user-oriented interpretations, making it difficult to assess and highlight policy properties quantitatively.

In addition to automata-based and program synthesis approaches, the advances in ILP \cite{cropper2022inductive} have also contributed to the development of explainable, logic-based frameworks for decision-making agents. ILP has already been employed to learn first-order theories from examples and background knowledge in the context of decision-making agents \cite{furelos2021induction,meli2024learning}, providing interpretable and causal explanations \cite{zhang2024critical,veronese2023inductive}. 
Furthermore, differentiable ILP ($\partial$ILP) \cite{evans2018learning} introduces gradient-based learning into ILP, enabling the integration of symbolic reasoning with neural perception models while maintaining interpretability. 
Overall, these logic-based methods offer rich semantic explanations and support formal guarantees over agent behavior. However, explainability in these approaches is assessed through user studies or readability evaluations, which tend to be subjective and inconsistent across different user groups \cite{ras2022explainable}.

\subsection{Open Challenges and Contribution}
In the above body of research, there is still limited attention to objective and user-independent evaluation criteria for XRL, addressing and quantifying crucial policy properties for trustworthy deployment in the real world, e.g., emerging behaviors and the robustness of the learned policy during training, the generalization capability of the extracted representations out of the training distribution, and the agreement and convergence among multiple agents in MARL \cite{retzlaff2024human,ras2022explainable}.

Our work aims to address these gaps by leveraging ILP-learned symbolic (logical) approximations of the RL policy and defining objective metrics for assessing the aforementioned properties.
Specifically, we focus on logical policy representations for their advantage in interpretable and scalable NeSy RL \cite{mao2019neuro,rocktaschel2017end,veronese2025online}, and for the high-level expressiveness which increases transparency and trustworthiness, e.g., in human-robot learning and interaction \cite{meli2025inductive}. More specifically, we choose the logical formalism of ASP, which represents the state of the art in the symbolic representation of planning domains \cite{meli2023logic}.
On top of the ASP approximations learned from sample policy executions, we define novel metrics to highlight different aspects of the RL training and generalization process.
These metrics include the \emph{activation rate} of the learned ASP policy representation, that is, how much the symbolic rules correctly reflect the actual behavior of the agent (useful to highlight which action-specific policies the agent is more confident on); \emph{feature coverage}, to quantify the impact of symbolic features in the decision-making process of the agent (similar to feature importance from XML); \emph{syntactic} and \emph{semantic distance} to measure how two symbolic (logical) policy approximations differ between each other, thus highlighting potential evolution of the skills of the agent during training or convergence between multiple agents.

% While the Co-12 \cite{nauta2023anecdotal} and LEAF \cite{amparore2021trust} frameworks represent major steps toward systematic evaluation of explainability, they are primarily designed for feature-attribution explanations in static supervised learning settings, thus resulting inapplicable in the RL context. In contrast, our work addresses symbolic, logic-based explanations extracted from RL agents and introduces evaluation metrics tailored to the structure and behavior of these representations. 

To the best of our knowledge, this is the first evaluation framework specifically focused on quantitatively explaining multiple aspects of RL and MARL policies.
Our metrics capture both the semantic and syntactic properties of symbolic policies and uniquely account for their temporal evolution across training. In this way, we directly address several critical limitations highlighted in the literature.
For instance, while \cite{darwiche2022quantifying} recognizes the potential of logical formalisms to explain black-box behaviors, it remains limited to \emph{feature relevance} and Boolean feature coverage, without exploring broader explanatory capabilities, such as policy-level insights, enabled by our proposed metrics.
A complementary approach is proposed by \cite{caporese2026RLmeetsLP}, where Logic Programming is employed to learn the environment's causal structure.
Moreover, \cite{retzlaff2024human} underscores the importance of explainability across different stages of the RL pipeline, particularly in human-centric and safety-critical applications. It highlights key desiderata such as understanding \emph{emergent behaviors} (e.g., inter-agent agreements in MARL) and assessing policy robustness under distributional shift. However, it also acknowledges the absence of objective, human-independent metrics to evaluate such properties, since most XRL work relies on subjective definitions of explainability. In contrast, our framework proposes concrete, quantitative metrics that make these aspects observable and measurable in an objective way, thus enabling principled assessments of explainability that go beyond feature importance and user perception.

\section{Background}\label{sec:background}

This section introduces the core concepts and tools relevant to our approach. 
We begin providing a formal overview of Markov Decision Processes (MDPs), the standard mathematical framework for modeling sequential decision-making problems under uncertainty \cite{SuttonBarto2018}. MDPs serve as the foundation for RL algorithms, which aim to compute optimal or near-optimal policies through interaction with an environment.
We then introduce the MDP formalization of the \emph{Intersection} domain, which will serve as a running example throughout the next sections. 
Then, we introduce Answer Set Programming (ASP) \cite{gelfond1988stable}, a state-of-the-art declarative logic programming paradigm used to represent planning domains \cite{meli2023logic}. 
Consequently, we present the general ILP \cite{Muggleton1991InductiveLogicProgramming} problem and its specialization to ASP via ILASP \cite{law2015learning}.

\subsection{Markov Decision Processes}

We model the agent's decision-making process using the formalism of a finite-horizon \textit{Markov Decision Process} (MDP), defined as a tuple $\mathcal{M} = \langle S, A, R, T, \gamma \rangle$, where:
\begin{itemize}
    \item \( S \) is a finite set of states;
    \item \( A \) is a finite set of actions;
    \item \( R: S \times A \rightarrow \mathbb{R} \) is the reward function;
    \item \( T: S \times A \times S \rightarrow [0,1] \) is the transition probability function, where \( T(s,a,s') = \Pr(s' \mid s,a) \);
    \item \( \gamma \in [0,1) \) is the discount factor.
\end{itemize}

The agent's objective is to find a policy \( \pi: S \rightarrow A \) that maximizes the expected cumulative reward, or return:
\begin{equation}
 \mathbb{E} \left[ \sum_{t=0}^{\infty} \gamma^t R(s_t, a_t) \right].\label{eq:ret}   
\end{equation}

Domains in which multiple agents act can be formalized as \textit{Multi-Agent Markov Decision Processes} (MAMDP), defined as a tuple $\mathcal{M}_n = \langle S, A^n, n, R, T, \gamma \rangle$, where:
\begin{itemize}
\item $S$ is a finite set of states of the environment, representing all agents;
\item $A^n = A \times \dotsb \times A$ (n times) is the product set of all finite action spaces, one for each agent; 
\item $n$ is the number of agents;
\item \( R: S \times A^n \rightarrow \mathbb{R} \) is the reward function;
\item \( T: S \times A^n \times S \rightarrow [0,1] \) is the transition probability function, where \( T(s,a,s') = \Pr(s' \mid s,a) \);
\item \( \gamma \in [0,1) \) is the discount factor.
\end{itemize}

If the whole MDP state is not observable by the agent, we define the task as a Partially Observable Markov Decision Process (POMDP). A POMDP is a tuple $\langle S, A, Z, O, R, T, \gamma \rangle$, with the following differences with respect to the previous MDP definition:
\begin{itemize}
    \item $Z$ denotes a finite set of \emph{observations}, that is, what the agent perceives about the state;
    \item $O: S \times A \rightarrow \Pi(Z)$ operates as the observation model;
    \item $T: S \times A \rightarrow \Pi(S)$ functions as the state-transition model with probability distribution $\Pi(S)$, also known as \emph{belief}, over states;
\end{itemize}  

Lastly, multiple agents in a partially observable environment are formalized as a Multi-Agent POMDP (MAPOMDP) $\langle S, A^n, n, Z, O, R, T, \gamma \rangle$, which combines the two previous definitions, dealing with observations and observation models, but also product sets of actions, one set for each agent.

The agent's objective is to compute a policy $\pi$: $\Pi(S) \rightarrow A$ that maximizes the discounted return, as already defined in Equation \ref{eq:ret}.

RL provides a data-driven approach for solving MDPs (and variants) when the transition dynamics and reward function are unknown. Through interaction with the environment, an RL agent learns to improve its policy over time. While RL has demonstrated impressive performance across various domains, its decision-making process is completely black-box. This lack of transparency has motivated growing interest in XRL, which aims to make learned policies and value functions interpretable and trustworthy.

\subsection{Running Example}

In the Intersection domain \cite{highway-env}, an agent (an autonomous vehicle) must navigate through a road intersection while avoiding collisions with other vehicles approaching from different directions. The environment simulates traffic dynamics where cars follow predefined lanes and movement patterns, but stochasticity in their behavior introduces uncertainty in the agent’s decision-making. The task requires the agent to balance safety and efficiency, deciding whether to accelerate, decelerate, or maintain constant velocity to cross the intersection successfully.

We formalize the Intersection domain as a Markov Decision Process (MDP). 
The state space $S$ consists of kinematic observations of up to fifteen vehicles. 
Each vehicle is represented by the features 
$\{x, y, v_x, v_y, \cos(h), \sin(h)\}$,
where the positions $(x,y)$ are bounded within $[-100,100]$, and the velocities $(v_x,v_y)$ within $[-20,20]$. The values of $\cos(h)$ and $\sin(h)$ represent the trigonometric heading of each vehicle.
These features provide the ego vehicle (i.e., the agent-controlled vehicle) with information about surrounding traffic dynamics in the intersection.  
The action space $A$ is discrete and defined by three possible longitudinal maneuvers: deceleration, maintaining constant speed, and acceleration.  
The reward function $R$ is defined to encourage safe and efficient driving. 
The agent receives a penalty of $-5$ in the event of a collision, a positive reward when maintaining a high speed, and a reward of $1$ upon successful arrival at the destination.  
The transition function $T$ is deterministic with respect to the ego vehicle’s actions, since each discrete action deterministically updates the vehicle’s kinematics. 
However, the behaviour of the other autonomous vehicles is decentralized and stochastic from the agent’s perspective, making the environment partially unpredictable.  

\begin{figure}[t]
    \centering
    \begin{subfigure}{0.57\linewidth}
        \centering
        \includegraphics[width=\linewidth]{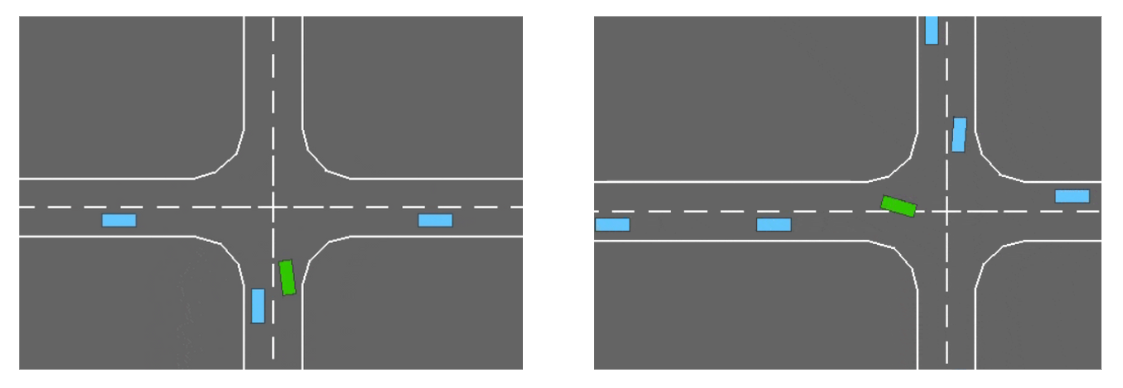}
        \caption{Example scenarios from the Intersection domain \cite{highway-env}.}
        \label{fig:highway}
    \end{subfigure}
    \hfill
    \begin{subfigure}{0.4\linewidth}
        \centering
        \includegraphics[width=\linewidth]{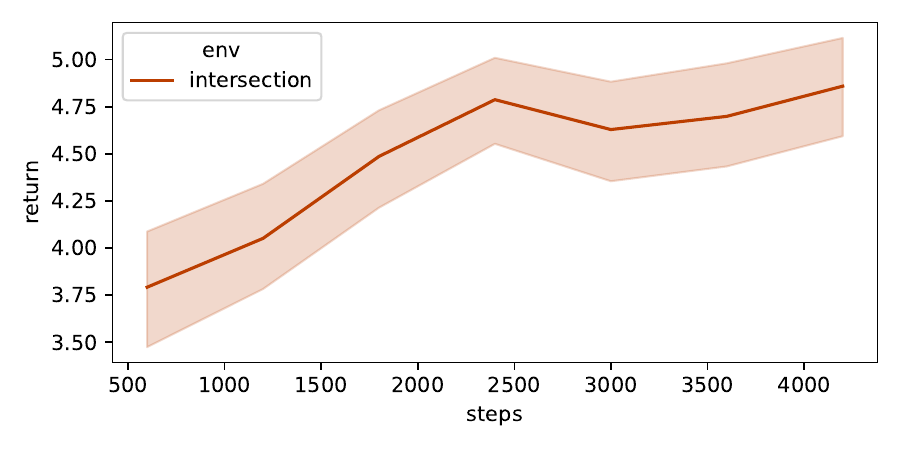}
        \caption{DQN training in the Intersection scenario.}
        \label{fig:highway_training}
    \end{subfigure}
    \caption{Visualization of the Intersection domain and corresponding training dynamics.}
    \label{fig:highway_combined}
\end{figure}

We select Intersection as a testing domain because, despite its relatively simple dynamics, it poses multiple challenges. These arise from the presence of multiple interacting agents, the stochastic behavior of other vehicles, and the need for long-horizon planning to ensure both collision avoidance and timely goal completion. We train a DQN \cite{silverDQN} agent to solve the task and extract logical policy representations from evaluation episodes extracted from different batches across the whole training. Figure \ref{fig:highway_training} shows the resulting training curve.

\subsection{Answer Set Programming}\label{sec:asp}

ASP is a declarative logic programming paradigm, particularly suited to represent complex domains and reason over them. An ASP program $P$ defines a domain as a set of logical statements, which encode relationships between entities in the form of predicates and variables (\emph{atoms}) \cite{gelfond1988stable}.

The core structure considered in this work is the \emph{normal rule}: 
\begin{equation}
    h \,\, \text{:-}\, \mathcal{B}, \,\text{with}\,\, \mathcal{B} = b_1, b_2, \ldots, b_n, \label{eq:normal_axiom}
\end{equation}
where $h$ is known as the \emph{head} of the rule and $\mathcal{B}$ forms the \emph{body}, a logical conjunction of literals (i.e., atoms or their negation) that must hold for the head to be derived. In logical terms, this rule corresponds to
$b_1 \wedge b_2 \wedge \dots \wedge b_n \rightarrow h$.

Variables in atoms are \emph{grounded} when they are assigned specific values from the domain. An atom is considered \emph{ground} if all its variables are grounded. The \emph{Herbrand base} of a program $P$, denoted $\mathcal{H}(P)$, is the set of all ground atoms that can be constructed using the predicate symbols and constants in $P$. 

ASP solvers compute the \emph{answer sets} of a program, that is, minimal models that satisfy all rules under the stable model semantics. Starting from an initial variable assignment, all ground atoms in the body of rules are computed; hence, ground head atoms
are derived. These models represent consistent, closed-world interpretations of the domain.

In the context of this work, we rely on body atoms to encode symbolic environmental features that describe the MDP state space $S$ (e.g., \stt{same\_lane(Id)} denoting the presence of a vehicle, identified by \stt{Id}, in the same lane of the ego vehicle, while head atoms correspond to available actions in the action space $A$ (e.g. \stt{faster} and \stt{slower} to increase and decrease agent's velocity).

\subsection{Inductive Logic Programming}

Inductive Logic Programming (ILP) is a symbolic machine learning framework that induces logical rules from structured data using a background theory and a formal hypothesis space. A generic ILP task under a logical formalism $F$ is defined as the tuple:
$\mathcal{T} = \langle B, S_M, E \rangle$,
where $B$ is the \emph{background knowledge}, a set of known logical statements (e.g., type constraints or static facts), $S_M$ is the \emph{search space}, i.e., the set of all rules expressible in $F$ that conform to a mode declaration $M$ \cite{muggleton1995inverse}, and $E = \langle E^+, E^- \rangle$ is the set of \emph{examples} to be covered by a learned hypothesis $H \subseteq S_M$. Specifically, the objective is to find a hypothesis $H$ such that $B \cup H \models E^+$ and $B \cup H \cup E^- \models \bot$, where $\models$ denotes logical entailment \cite{Muggleton1991InductiveLogicProgramming}. We refer to $E^+$ as the set of \emph{positive examples}, and to $E^-$ as the set of \emph{negative examples}.
In other words, the goal of the learning task is to find a subset $H$ of the search space $S_M$ that, when combined with background knowledge $B$, entails the positive examples and does not entail the negative examples. 

In this work, we consider ILP under the ASP semantics.  Each example $e \in E$ is a Context-Dependent Partial Interpretation (CDPI), defined as a tuple $\langle e, C \rangle$, where $e = \langle e^{inc}, e^{exc} \rangle$ is a partial interpretation, with $e^{inc} \subseteq \mathcal{H}(P)$ and $e^{exc} \subseteq \mathcal{H}(P)$ representing, respectively, the atoms that must and must not be present in the answer set, and $C \subseteq \mathcal{H(F)}$ is the \emph{context}, a set of ground atoms describing the specific scenario \cite{law2015learning}. For the aim of this work, when building the set $E$, we only consider positive examples $E^+$, since we will retrieve them only from observed state-action pairs in RL execution traces.
The goal of ILP is to find a hypothesis $H$ such that for every example $e = \langle \langle e^{inc}, e^{exc} \rangle, C \rangle \in E$, there exists at least one answer set $as \in AS(B \cup H \cup C)$ satisfying:
\begin{equation}\label{eq:ilasp}
\forall e \in E \ \exists as \in AS(B \cup H \cup C) \text{ such that } e^{inc} \subseteq as,\ e^{exc} \cap as = \emptyset.
\end{equation}
To this aim, we employ the ILASP system \cite{law2015learning}, which is designed to induce the hypothesis under the ASP semantics. The goal of ILASP is to find the \emph{shortest} $H$ (i.e., with the minimal number of atoms for easier interpretability). We use $|H|$ to refer to the length of hypothesis $H$.
In our context, examples are extracted from MDP execution traces, i.e. sequences of state-action pairs.
In this way, finding $H$ results in obtaining a symbolic representation of the policy followed by the RL agent, expressed in a human-interpretable logical form. This symbolic encoding enhances explainability by allowing practitioners to inspect, validate, and reason about the agent's decision-making process in terms of explicit rules derived from its behavior.
\section{Methodology}\label{sec:methodology}
\begin{figure}
    \centering
    \includegraphics[width=0.8\linewidth]{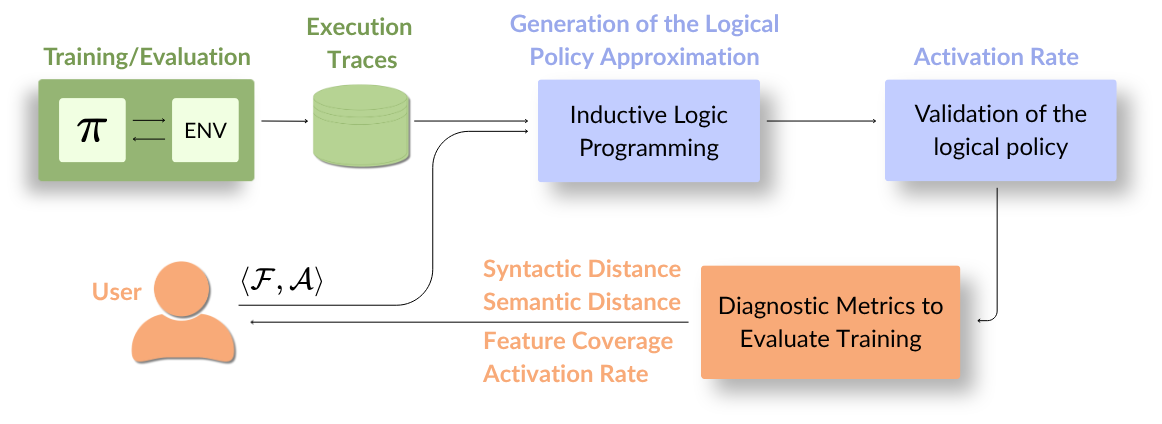}
    \caption{Overview of the proposed framework.}
    \label{fig:flowchart}
\end{figure}

In this section, we describe the process by which we generate a logical policy approximation starting from the observation of RL execution traces, enabling symbolic interpretability and the application of our explainability metrics.
Figure \ref{fig:flowchart} shows an overview of the proposed framework.
The input to this pipeline consists of a set of RL execution traces, in the form of state-action pairs, and a formalization of the task domain, defined as a pair $\langle \mathcal{F}, \mathcal{A} \rangle$, where $\mathcal{F}$ denotes the set of fluents describing the environment and $\mathcal{A}$ the set of available actions. The output of the ILP task is a hypothesis $H$, representing a logical policy approximation of the neural policy $\pi$ followed by the RL agent in the provided execution traces, expressed in the ASP formalism. We then apply our explainability metrics on $H$ to retrieve relevant information about the agents' training process and the generalization and robustness of the policy.

In the following, we explain how we translate state-action trajectories generated from $\pi$ into a logical representation consistent with the chosen task formalization $\langle\mathcal{F},\mathcal{A}\rangle$. We then show how this formalization is leveraged to construct a learning task for ILASP \cite{law2015learning}, the ILP specialization to the ASP semantics, which synthesizes a symbolic policy approximation $H$ that captures the agent's behavior in a human-interpretable form. Then, we formalize and present a set of novel objective explainability metrics for XRL. For each step of the presented approach, we include a running example from the \emph{Intersection} task of the \emph{highway-env} environment \cite{highway-env}.

\subsection{ASP Formalization of the MDP Domain}
We begin by representing the MDP domain using the logical formalism of ASP. 
To bridge the state and action spaces of the MDP with the corresponding symbolic representation $\langle\mathcal{F,A}\rangle$, we need to design a \emph{feature map} $F_\mathcal{F} : S \rightarrow \mathcal{H(F)}$ and an \emph{action map} $F_\mathcal{A} : A \rightarrow \mathcal{H(A)}$, with $\mathcal{H(F)}$ and $\mathcal{H(A)}$ denoting the Herbrand bases of $\mathcal{F}$ and $\mathcal{A}$, respectively. 
More specifically, the feature map $F_\mathcal{F}$ translates an MDP state $s \in S$ into a logical description composed of ground terms from $\mathcal{F}$ (e.g., \stt{same\_lane(id)} denoting the presence of another vehicle, identified by \stt{id}, on the same lane of the ego agent, in Intersection). Similarly, the action map $F_\mathcal{A}$ provides a logical representation of MDP actions using ground terms from $\mathcal{H(A)}$. We define $F_\mathcal{A}$ as a straightforward symbolic encoding of the MDP's action space, mapping each action $a \in A$ to a corresponding term in $\mathcal{H(A)}$ (e.g., \stt{idle}, \stt{faster} and \stt{slower} in Intersection).
This formalization process relies on the following simple assumptions.
\begin{assumption}
\label{ass:s_o} 
We design $F_\mathcal{F} : S_o \subseteq S \rightarrow \mathcal{H(F)}$, such that $S_o$ is the observable sub-set of MDP states $S$. 
\end{assumption}
This implies that, in partially observable contexts, we define $F_\mathcal{F}$ over $Z$ instead of $S$.
This allows us to maintain generality across different problem classes, while focusing the symbolic representation on features relevant for decision-making and directly available to the agent. 
\begin{assumption}
\label{ass:a} 
$\mathcal{A}$ directly encodes MDP actions. 
\end{assumption} 
This assumption serves as a bridge between the continuous nature of RL action spaces and the discrete symbolic format required by ILASP.

For the Intersection domain, action atoms are constructed as a one-to-one mapping of the MDP actions' set $A$, resulting in $\mathcal{A} = \{\stt{idle},\stt{slower}, \stt{faster}\}$. The feature set $\mathcal{F}$ contains the following predicates: \stt{obs\_is\_far(Id,Dist)} and \stt{obs\_is\_close(Id,Dist)}, stating that a vehicle (identified by \stt{Id}) is at a distance greater (resp. lower) than \stt{Dist} from the ego agent, \stt{obs\_is\_crossing(Id)} and  \stt{ego\_is\_crossing}, indicating that a vehicle (or the ego agent) is currently crossing the intersection, \stt{same\_lane(Id)}, to keep track of the vehicles occupying the same lane of the ego agent, and \stt{obs\_front(Id)} for the ones that are directly in front of it. The range of the \stt{Id} variable is $[0,14]$, since the environment comprises $15$ agents, while we set \stt{Dist} $\in [0,100]$.
The only information needed to build $\mathcal{F}$ is the positions of the agent and the other vehicles in the environment, all of which are included in $S$.

\subsection{Generation of Logical Policy Approximations via ILP}
Given a set of RL execution traces $T = \{t_1, \dots, t_N\}$, where each trace $t_i$ is a sequence of state-action pairs $ \langle s, a \rangle $ with $s\in S$ and $a\in A$, our goal is to derive a logical policy approximation by formulating an ILP task $\mathcal{T} = \langle B, S_M, E \rangle$. This task can be solved using ILASP to synthesize a symbolic policy that captures the observed behavior of a trained RL agent.

To construct the ILASP task, we encode each state-action pair $ \langle s, a \rangle $ into a CDPI $\langle \langle e^{inc}, e^{exc}\rangle, C\rangle$, with $C = F_\mathcal{F}(s)$, $e^{inc} = F_\mathcal{A}(a)$ and $e^{exc} = \{F_\mathcal{A}(a')\, |\; a' \in A \setminus a \}$. 
That is, we populate the context $C$ with the state observed by the RL agent, the included set $e^{inc}$ with the action accomplished by the agent in reaction to that state, and the excluded set $e^{exc}$ with all the other actions available to the agent but not observed at that specific timestep.

The background knowledge $B$ of the task only contains the definition of the ASP variables and ranges, and in $S_M$ we only consider normal rules, as defined in Equation \ref{eq:normal_axiom}, with head $h$ and body $\mathcal{B}$ composed by atoms $f_1, \ldots, f_n \in \mathcal{F} \cup \{X \gtrless x\}$, with $X$ being any variable and $x$ an integer constant. 

Once the task $\mathcal{T}$ is defined, we invoke ILASP to infer a hypothesis $H\subseteq S_M$ that satisfies Equation \ref{eq:ilasp}. The result is a compact and human-readable logical policy approximation of the original black-box policy $\pi$, allowing us to objectively evaluate it through the metrics defined in the following. 

In the Intersection scenario, for example, the logical policy approximation $H$ learned from the DQN agent at the convergence of the training process (4000 execution steps in Figure \ref{fig:highway_training}) is the following:

\begin{align}
    &\stt{slower :- obs\_is\_crossing(V1); obs\_is\_close(V1, 35).}\\
    &\stt{idle :- obs\_is\_crossing(V1); obs\_is\_close(V1, 55); obs\_is\_far(V1, 50).}\\
    &\stt{faster :- not obs\_is\_crossing(V1); obs\_is\_far(V1, 45).}
\end{align}
Meaning that the ego agent regulates its speed mainly depending on how far it is with respect to the other vehicles crossing the intersection, only going faster if no other vehicle is on the way. 

\subsection{Explainability Metrics for Evaluating Logical Policy Approximations}\label{sec:metrics}
We now define a set of explainability metrics designed to evaluate the RL policy, given the symbolic approximation induced via ILASP. These metrics quantify different aspects of the RL agent's policy and aim to provide useful insights beyond the return of the agent, e.g., the actual convergence of the sub-policies of specific actions, the robustness of the learned behaviour across training, the impact of user-defined environmental features $\mathcal{F}$, the generalization capabilities of the policy, and the stationarity and agreement in MARL. 
Specifically, we formalize the following metrics: \emph{activation rate}, \emph{feature coverage}, \emph{syntactic distance}, and \emph{semantic distance}.
We also provide an overview of how these metrics can be used to evaluate the RL training process.
To understand the following definitions, recall that, given a logical policy approximation $H$, we use $AS_H(F_\mathcal{F}(s))$ to refer to the answer set of $H$ given the context grounded from the MDP state $s$ using the map $F_\mathcal{F}$.

\subsubsection{Activation rate}
This metric quantifies the frequency with which each rule contributes to decision-making, highlighting both the relevance and precision of specific rules within the logical policy approximations.
\begin{definition}
Let $H$ be a logical policy approximation, and $T = \{t_1, t_2, \dots, t_N\}$ a set of $N$ execution traces collected from an RL agent. 
The \textbf{activation rate} for a rule $r_k \in H$ with head $h_k$ and body $\mathcal{B}_k$, as for equation \ref{eq:normal_axiom}, is computed as:
\begin{equation}
\alpha(r_k) = \frac{\sum_{i=1}^N\sum_{j=1}^n \mathbbm{1} \{ F_\mathcal{A}(a^i_j) \in AS_{r_k}(F_\mathcal{F}(s^i_j))\}}{\sum_{i=1}^N\sum_{j=1}^n \mathbbm 1 \{F_\mathcal{A}(a^i_j) \in \mathcal{H}(h_k)\}}\label{eq:activation_rate_rule}
\end{equation}
where $\mathbbm 1\{A\} = 1$ if $A$ is true, 0 otherwise.
\end{definition}
Intuitively, each time the agent performs the action represented by the head $h_k$ of the rule, we check whether the state observed by the agent implies that specific action, according to the body of the rule. We then normalise the obtained value, dividing it by the number of times that same action has been performed by the RL agent.
Considering that a theory could include multiple rules for the same action (i.e., more than one rule with the same head), we also define the activation rate for each action $\mathscr{a} \in \mathcal{A}$ as follows:
\begin{equation}
\alpha(\mathscr{a}) = \frac{\sum_{i=1}^N\sum_{j=1}^n \mathbbm{1} \{ F_\mathcal{A}(a^i_j) \in \mathcal{H}(\mathscr{a}) \land F_\mathcal{A}(a^i_j) \in AS_H(F_\mathcal{F}(s^i_j)) \}}{\sum_{i=1}^N\sum_{j=1}^n \mathbbm 1 \{F_\mathcal{A}(a^i_j) \in \mathcal{H}(\mathscr{a})\}}
\label{eq:activation_rate_action}
\end{equation} 
The activation rate of the whole theory $\alpha(H)$ can be easily computed either as the sum of the activation rate for each action $\mathscr{a} \in \mathcal{A}$ or as the sum over $k$ of each $\alpha(r_k)$.

%The activation rate provides an objective measure of how frequently the symbolic representation of the policy aligns with the behavior exhibited by the agent from which it has been extracted, in a fine-grained action level. 
The activation rate provides an objective measure not only of how frequently the symbolic representation of the policy as a whole aligns with the agent’s behavior, but more importantly of which specific parts of the policy, down to the level of individual actions, are actually well represented. In this sense, the activation rate offers more fine-grained insights that go beyond what can be captured by the global return alone.
It supports trust and interpretability by identifying which rules are actively guiding decisions.  
Furthermore, tracking the activation rate across different checkpoints along the RL training curve offers insights into the temporal evolution of the policy during training. It can reveal whether the agent is converging towards stable decision patterns or undergoing significant shifts that may warrant further investigation. 

For instance, consider Figure \ref{highway_activation_rate}, in which the activation rate $\alpha(\mathscr{a})$ for each action in the Intersection domain is visualized. We see that the activation rate is particularly high for action \stt{slower} and \stt{faster}, while it remains lower for action \stt{idle}. This suggests that the action \stt{idle} is more difficult to capture in logical terms, since it is executed in a wide variety of contexts (e.g., maintaining a low constant speed is quite different from maintaining a high constant speed). In contrast, the actions \stt{slower} and \stt{faster} tend to be performed according to more precise and well–defined strategies, which results in a higher and more stable activation rate.

As we will show in the experimental section, the activation rate is particularly valuable for identifying overfitting on specific actions/skills, instability, or abrupt changes in the policy behavior, e.g., due to the non-stationarity in MARL \cite{foerster2016learning}.
Moreover, the activation rate can also be computed in out-of-distribution scenarios to assess the generalization and transferability of symbolic rules in different settings than training.  

Additionally, the frequency of rule activation may serve as a proxy for rule importance, supporting the prioritization of explanations in post hoc analysis or during human-AI interaction (e.g. a rule that is rarely activated may indicate redundancy or over-specificity). In interactive or human-in-the-loop systems, this metric can inform the selective presentation of rules, ensuring that the most influential ones are communicated to users, thus enhancing transparency and trust \cite{amparore2021trust}.

\subsubsection{Feature coverage}
Feature coverage aims at measuring how relevant a feature is with respect to the whole policy representation. 
\begin{definition}
Given the set of all features $\mathcal{F}$ available in the domain, let $\mathcal{F}_H \subseteq \mathcal{F}$ be the subset of symbolic features used in the body of the rules in $H$. For each feature $\mathscr{f} \in \mathcal{F}_H$, \textbf{feature coverage} is computed as:
\begin{equation}
   \phi(\mathscr{f}) = \frac{\sum_{k=1}^{|H|} \sum_{b^k_m \in \mathcal{B}_k} \mathbbm 1 \{b^k_m = \mathscr{f} \}}{\sum_{k=1}^{|H|} |b^k|} \label{eq:feature_coverage}
\end{equation}
with $\mathcal{B}_k= \bigwedge_{m=1}^M b^k_m$ indicating the body of $k$\textsuperscript{th} rule, and $|H|$ the number of rules composing $H$.  
\end{definition}
Namely, feature coverage quantifies the proportion of the symbolic theory that involves a specific feature $\mathscr{f}$, offering a measure of how prominently that feature contributes to explaining the behavior of the agent. 
% In this way, we obtain an estimate of the relative importance of different aspects of the domain.

In terms of explainability, feature coverage helps identify which parts of the input space are most influential in driving agent behavior. Features with high coverage can be considered critical for decision-making, whereas those with minimal presence may be deemed less relevant, potentially suggesting redundancy or poor feature design. By identifying underutilized or overly dominant features, one can adjust the feature set to enhance the symbolic model's fidelity and interoperability \cite{perotti2024metrics}. This can be especially beneficial in iterative, human-in-the-loop learning pipelines, where domain experts aim to guide or audit the learning process \cite{retzlaff2024human}.

\begin{figure}
    \centering
    \begin{subfigure}[t]{0.49\linewidth}
        \centering
        \includegraphics[width=\linewidth]{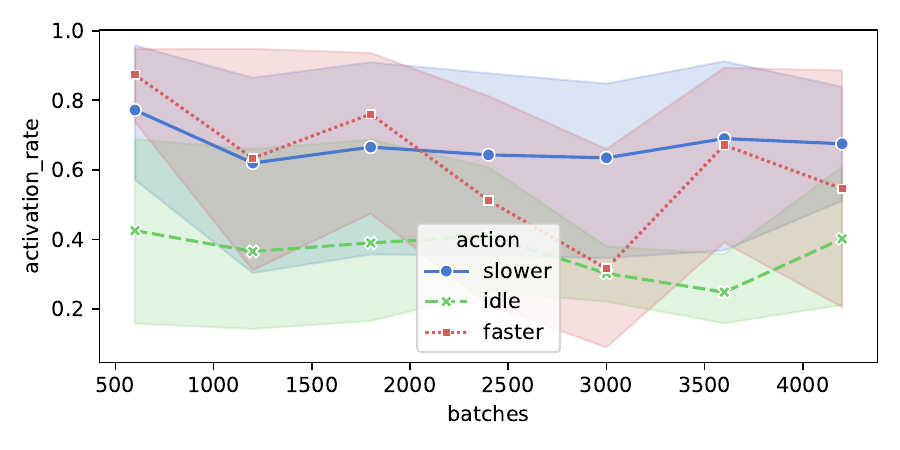}
        \caption{Activation rate $\alpha(\mathscr{a})$.}\label{highway_activation_rate}
    \end{subfigure}
    \hfill
    \begin{subfigure}[t]{0.49\linewidth}
        \centering
        \includegraphics[width=\linewidth]{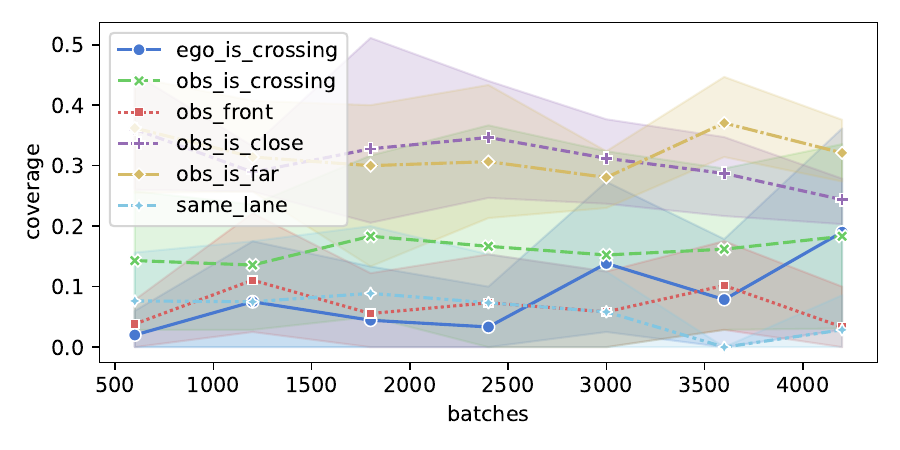}
        \caption{Feature coverage $\phi(\mathscr{f})$.}\label{highway_feature_cov}
    \end{subfigure}

    \caption{Activation rate and feature coverage on the Intersection domain, computed on logical policy approximations learned for each batch (100 episodes each) during training of a simple tabular RL agent.}
    \label{fig:intersec_full_analysis_act_feat}
\end{figure}

Figure \ref{highway_feature_cov}, for example, shows that in the policy approximations induced from the RL agent trained to solve the Intersection task, the distance from other vehicles (represented by the \stt{obs\_is\_far} and \stt{obs\_is\_close} predicates), together with the presence of vehicles crossing the intersection, is the most valuable thing for the ego agent in order to decide when to change the velocity. On the contrary, the \stt{obs\_front} and \stt{same\_lane} predicates seem to be almost unnecessary to the representation of the agent's policy.

From a comparative standpoint, feature coverage assesses how consistently certain features are important across the training process of the agent, or for different agents in MARL settings. For instance, stable feature coverage patterns across training runs may signal the crucial impact of one feature from the early stages of training, while a large variance might indicate sensitivity to initial conditions or environmental changes.
Finally, feature coverage contributes to the goal of aligning learned models with human reasoning. By highlighting which domain concepts (as captured by features) are central to the agent’s logic, the metric supports more meaningful, human-centered explanations \cite{sequeira2020interestigness}.

\subsubsection{Syntactic distance}
Syntactic distance captures the structural dissimilarity between two symbolic theories, focusing on the differences in the specific rule predicates that are involved when explaining the agent's policy. This is particularly relevant to understand whether two policies agree not only according to their semantics, but also in terms of the specific atoms and domain concepts they rely on. Extending the preliminary formulation we introduced in \cite{veronese2025online}, we now give a formal definition:
\begin{definition}
Given two learned theories $H_1$ and $H_2$, and a specific action $\mathscr{a} \in \mathcal{A}$, we define the \textbf{syntactic distance} as follows:

\begin{equation}
\lambda(\mathscr{a}; H_1, H_2) = 1 - \frac{|\{\mathcal{B}_k | \, r_k \in H_1 \land h_k = \mathscr{a} \}\cap \{\mathcal{B}_k | \, r_k \in H_2 \land h_k = \mathscr{a} \}|}{|\{\mathcal{B}_k | \, r_k \in H_1 \land h_k = \mathscr{a} \}\cup \{\mathcal{B}_k | \, r_k \in H_2 \land h_k = \mathscr{a} \}|} \label{eq:syntactic_distance}
\end{equation}
where each rule $r_k$ is represented by its head $h_k$ and body $\mathcal{B}_k$, as defined in \ref{eq:normal_axiom}. 
\end{definition}
This metric computes the Jaccard index (i.e., the intersection over union) between the sets of rule bodies associated with the same action in each theory, quantifying the degree of overlap.
To obtain a global syntactic divergence, we extend this measure to the full action space, leveraging syntactic distance over all actions:
\begin{definition}
Let $H_1, H_2$ be two logical policy approximations. Given the syntactic distance $\lambda$, computed over each action $\mathscr{a} \in \mathcal{A}$, we compute the \textbf{average syntactic distance} between $H_1$ and $H_2$ as:
\[
\bar{\lambda}(\mathcal{A}; {H}_1, {H}_2) = \frac{1}{|\mathcal{A}|} \sum_{i=1}^{|\mathcal{A}|} \lambda(\mathscr{a}_i; {H}_1, {H}_2).
\]
\end{definition}

Syntactic distance offers insight into how two symbolic policies are constructed: a low value indicates that the policies rely on the same features to make decisions. Conversely, a high syntactic distance may suggest divergent reasoning paths. This makes the metric especially useful in evaluating interpretability and debugging learned rules. In particular, syntactic distance can be useful in scenarios involving explainability verification or expert auditing, where it is crucial to ensure that the learned policy adheres to known or desired reasoning patterns \cite{retzlaff2024human}.

Finally, the syntactic distance can also be employed to evaluate the consistency of an agent's policy over the training process. By comparing theories extracted from different batches during training (e.g., $H_1$ and $H_2$ corresponding to early and later checkpoints of the training of the same agent), it is possible to assess the degree of convergence of the symbolic policy and, consequently, of the training process. A decreasing syntactic distance across training iterations may, in fact, indicate stabilization and refinement of the learned policy, providing additional insights into the learning dynamics.
In the following, we will refer to this analysis as \emph{syntactic convergence}, as it will capture the structural similarity of the policies learned at subsequent batches of RL training.

\begin{figure}
    \centering
    \begin{subfigure}[t]{0.49\linewidth}
        \centering
        \includegraphics[width=\linewidth]{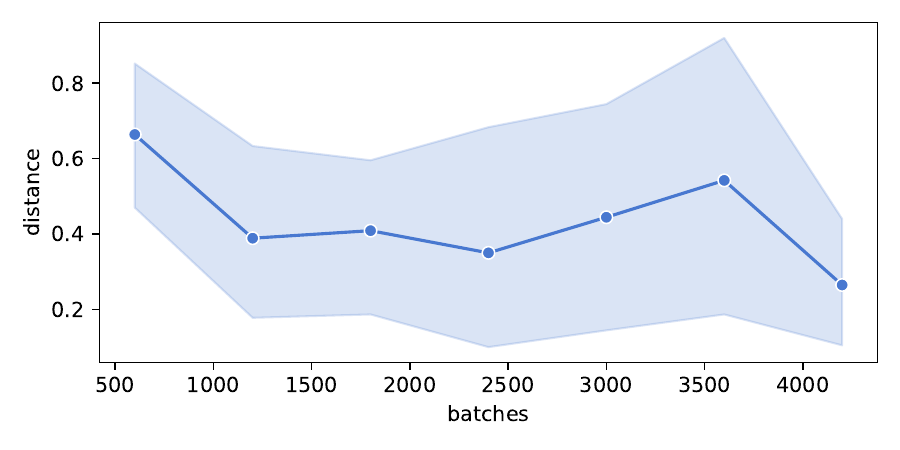}
        \caption{Syntactic convergence between theories of the single agent at different stages of the RL training.}\label{highway_synt_dist}
    \end{subfigure}
    \hfill
    \begin{subfigure}[t]{0.49\linewidth}
        \centering
        \includegraphics[width=\linewidth]{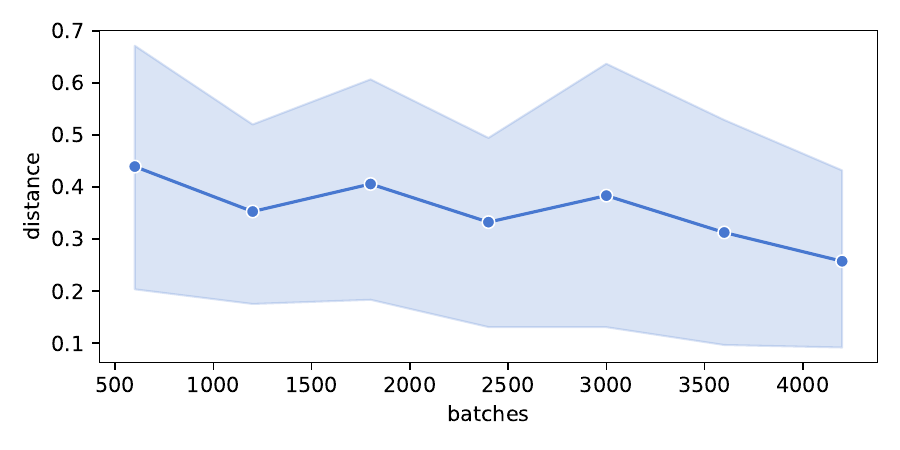}
        \caption{Semantic convergence between theories of the single agent at different points of the training.}\label{highway_sem_dist}
    \end{subfigure}

    \caption{Syntactic and semantic distances on the Intersection domain, computed on logical policy approximations learned for each batch (100 episodes each) during training of a simple tabular RL agent.}
    \label{fig:intersec_full_analysis_dist}
\end{figure}

For instance, Figure \ref{highway_synt_dist} shows the syntactic distance (over the full logical approximation) measured in the Intersection domain between representations extracted at each training batch and the final one at convergence.
% Specifically,  we compute both the distance between the symbolic approximations extracted from sequential training batches (blue) and between each batch and the final one at convergence (green).
The curve evidences that the logical theories learned across the whole training process remain syntactically distant, but slowly converge towards the final checkpoint, meaning that the policy followed by the agent reflects the convergence of the training curve, shown in Figure \ref{fig:highway_training}.

\subsubsection{Semantic distance}
The semantic distance measures the behavioral divergence between theories under identical conditions in terms of the state observed by the RL agent.
\begin{definition}
Given a fixed grounded context $c = F_\mathcal{F}(s)$, We define \textbf{semantic distance} between two logical policy approximations $H_1$ and $H_2$ as:
\begin{equation}
\delta(c; H_1, H_2) = 1 - \frac{|AS_{H_1}(c)\cap AS_{H_2}(c)|}{|AS_{H_1}(c)\cup AS_{H_2}(c)|}. 
\label{eq:semantic_distance}
\end{equation}
Over a set $C = \{c_1, \dots, c_m\}$\ of $m$ grounded contexts, we compute the \textbf{average semantic distance} as:
\begin{equation}
\bar{\delta}(C; {H}_1, {H}_2) = \frac{1}{m} \sum_{i=1}^{m} \delta(c_i; {H}_1, {H}_2).\label{eq:avg_semantic_dist}
\end{equation}
\end{definition}
That is, we measure the Jaccard index between the answer sets generated by $H_1$ and $H_2$ given the same set of grounded features.
Moreover, the semantic distance can be computed in a fine-grained, action-specific manner by restricting the comparison to the subsets of $H_1$ and $H_2$ whose rules have head predicates corresponding to a particular action $\mathscr{a}$:
\begin{equation}
\bar{\delta}(C; {H}^\mathscr{a}_1, {H}^\mathscr{a}_2) = \frac{1}{m} \sum_{i=1}^{m} \delta(c_i; {H}^\mathscr{a}_1, {H}^\mathscr{a}_2).\label{eq:avg_semantic_dist_action}
\end{equation}

This metric plays a pivotal role in explainability by offering a behavioral lens through which symbolic theories can be compared, beyond the mere structural convergence evidenced by the syntactic distance. A low semantic distance indicates that two theories (hence the corresponding policies) are behaviorally similar, producing consistent predictions across shared input conditions, even if their syntactic distance remains high. Conversely, high semantic divergence may point to fundamental differences in learned strategies, potentially revealing variation in training dynamics, feature utilization, or inductive biases.

The semantic distance is especially informative in scenarios involving multi-agent learning or ensemble methods, where comparing policies is essential for evaluating convergence, robustness, or diversity, especially in non-stationary contexts \cite{retzlaff2024human}. 

As with semantic distance, we can apply this metric to policies extracted from different stages of the same agent’s training process. In this way, we can evaluate \textit{semantic convergence}, namely, the semantic distance between a theory at a certain training step and the theory at convergence, to investigate the stability and convergence of symbolic policies over time.

Figure \ref{highway_sem_dist} clearly shows that, although the syntactic distance between approximations extracted from different batches in the Intersection domain tends to follow a pattern similar to that of the semantic distance, the two measures behave differently and convey complementary information. 
As for the syntactic distance, the semantic distance slowly converges with the training curve and beyond, but in a more stable way with respect to the syntactic distance. This behavior indicates that the logical interpretation of the agent’s strategy becomes more and more robust: even if small syntactic variations persist in the representation of the rules, these differences do not substantially alter the semantics of the policy. In other words, the agent continues to act according to a coherent decision logic, and the semantic perspective allows us to capture this stability more faithfully than the purely syntactic comparison.

\section{Experiments}\label{sec:experiments}

\begin{figure}
    \centering
    \begin{subfigure}[t]{0.4\linewidth}
        \centering
        \includegraphics[width=0.6\linewidth]{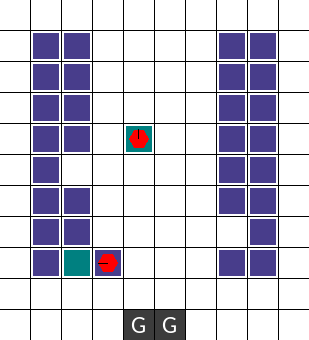}
        \caption{RWARE \citep{papoudakis2021benchmarking}}\label{fig:rware}
    \end{subfigure}
    \hspace{.5cm}
    \begin{subfigure}[t]{0.4\linewidth}
        \centering
        \includegraphics[width=0.6\linewidth]{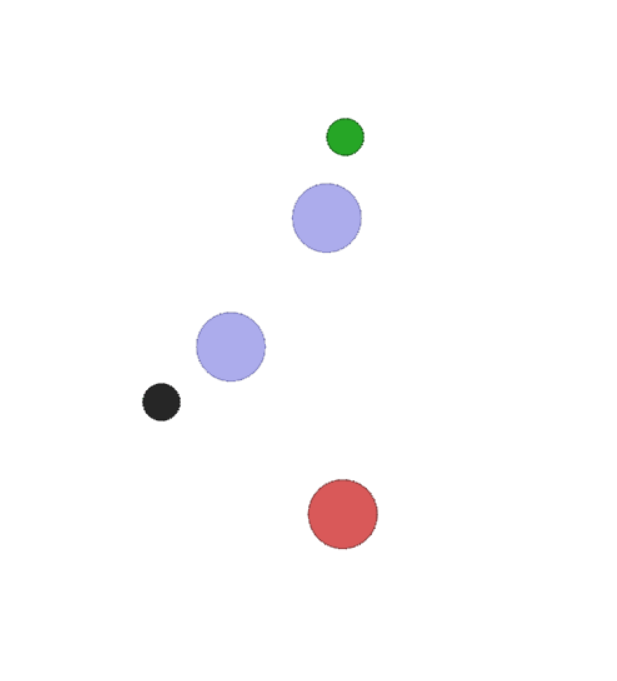}
        \caption{Simple Adversary \citep{lowe2017multi, mordatch2017emergence}}\label{fig:mpe}
    \end{subfigure}

    \begin{subfigure}[b]{0.49\linewidth}
        \centering
        \includegraphics[width=\linewidth]{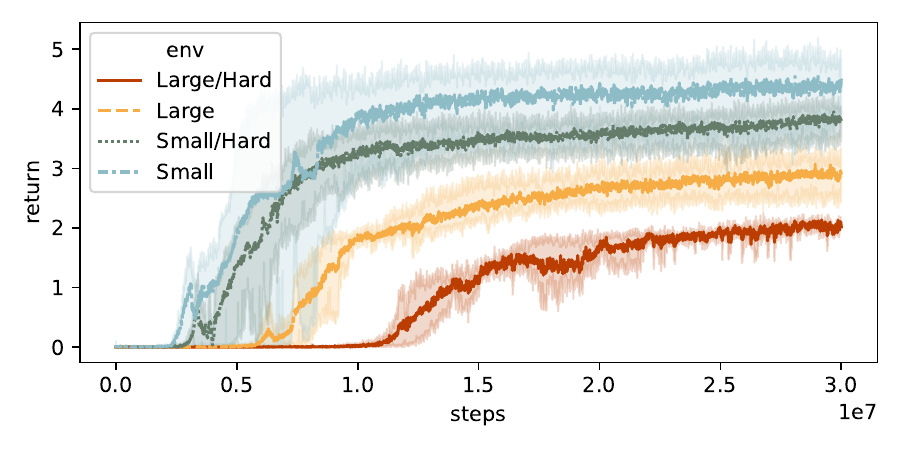}
        \caption{RWARE}
        \label{fig:training_rware}
    \end{subfigure}
    \hfill
    \begin{subfigure}[b]{0.49\linewidth}
        \centering
        \includegraphics[width=\linewidth]{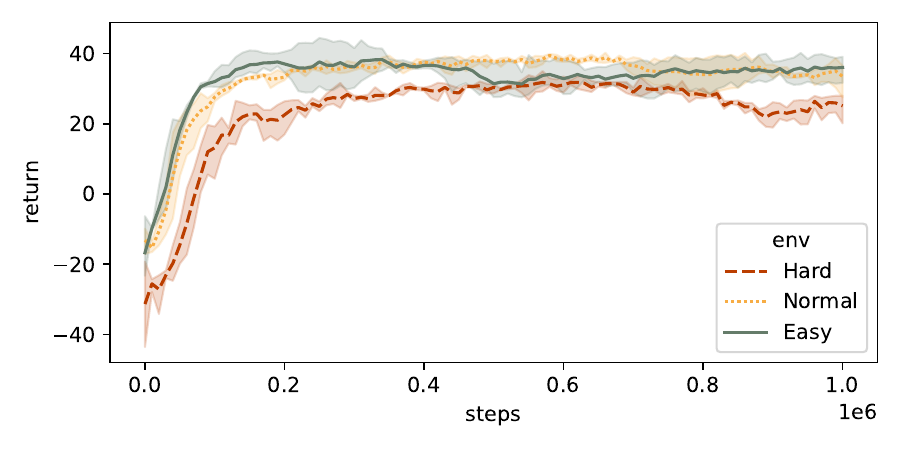}
        \caption{Simple Adversary}
        \label{fig:training_mpe}
    \end{subfigure}
    \caption{MARL Environments and training curves in different variants of the domains.  X-axis: training steps. Y-axis: average return (aggregated for all agents).}
    \label{fig:environments_and_training}
\end{figure}

In this section, we present our experimental evaluation on two MARL domains: \textit{RWARE} \citep{papoudakis2021benchmarking} (Figure \ref{fig:rware}, and \textit{Simple Adversary} \citep{lowe2017multi, mordatch2017emergence} (Figure \ref{fig:mpe}) \footnote{All the experiments shown in the following section and Appendix are available at \href{https://github.com/Isla-lab/explainability_metrics_for_rl}{https://github.com/Isla-lab/explainability\_metrics\_for\_rl}}. The nature of the two domains is different, in that RWARE is \textit{collaborative}, i.e., agents must coordinate to deliver the highest number of good shelves in a warehouse setting. Instead, Simple Adversary is \textit{contrastive}, i.e., two agents must defend a goal location. At the same time, another one must reach it, and the reward reflects how well they have performed their task separately. 

For each domain, we train a MAPPO \citep{Chao2021MAPPO} policy (Figures \ref{fig:training_rware} and \ref{fig:training_mpe}) using the \textit{epymarl} library \citep{papoudakis2021benchmarking}in relatively small domain instances.
We then learn logical theories associated with the agent at different checkpoints of training, and we use them to compute our explainability metrics, both in the same training environment and in more general environments. 

For each domain, we will address the following \textbf{research questions}:
\begin{itemize}
    \item [\RQinsighttraining)] Do the proposed metrics provide more insight into the training process of an RL agent than the classical return curve, e.g., by evaluating the robustness and convergence of sub-policies for specific actions (and related logical approximations)? 
    \item [\RQinsightmultiagent)] In the context of multi-agent domains, can we gain a deeper understanding of the differences and similarities between the policies learned by different agents by computing informative inter-agent explainability metrics over the symbolic approximations of these policies?
    \item [\RQfeatures)] Can feature importance indicate how determinant the pre-defined features are in the decision process, both in terms of the aggregate policy and sub-policies for the single actions?
    \item [\RQknowledge)] Can the proposed metrics be used to select symbolic knowledge to be later used for generalizable decision making, e.g., by identifying the most influential environmental features and actions in out-of-distribution scenarios with respect to the training setting?
\end{itemize}

The remainder of the section is organised as follows: first, we will formalise (both as an MDP and in the corresponding ASP representation) the two domains considered in our experiments, showing some examples of the learned logical policy approximations. We will then address each research question separately, with examples from both domains, showing plots to visualize the evolution of the metrics during the training process.

% \begin{figure}
%     \centering
%     \begin{subfigure}[t]{0.4\linewidth}
%         \centering
%         \includegraphics[width=0.6\linewidth]{images/rware/rware.png}
%         \caption{RWARE \citep{papoudakis2021benchmarking}}\label{fig:rware}
%     \end{subfigure}
%     \hspace{.5cm}
%     \begin{subfigure}[t]{0.4\linewidth}
%         \centering
%         \includegraphics[width=0.6\linewidth]{images/MPE/MPE.png}
%         \caption{Simple Adversary \citep{lowe2017multi, mordatch2017emergence}}\label{fig:mpe}
%     \end{subfigure}

%     \caption{MARL Environments.}
%     \label{fig:environments}
% \end{figure}

% \begin{figure}
%     \centering
%     \begin{subfigure}[b]{0.49\linewidth}
%         \centering
%         \includegraphics[width=\linewidth]{images/rware/runs2.pdf}
%         \caption{RWARE}
%         \label{fig:training_rware}
%     \end{subfigure}
%     \hfill
%     \begin{subfigure}[b]{0.49\linewidth}
%         \centering
%         \includegraphics[width=\linewidth]{images/MPE/runs.pdf}
%         \caption{Simple Adversary}
%         \label{fig:training_mpe}
%     \end{subfigure}
%     \caption{Training curves. X-axis: training steps. Y-axis: average return (aggregated for all agents).}
%     \label{fig:training}
% \end{figure}

\subsection{Collaborative Multi-Agent Domain: RWARE}\label{sec:rware_formalization}
RWARE \citep{papoudakis2021benchmarking} (Figure \ref{fig:rware}) is a challenging, partially-observable cooperative multi-agent MDP. In it, $n$ agents navigate a grid-like warehouse with $m$ shelves, of which $\ell$ are marked as \textit{good} (i.e., ready for delivery) at any time. Agents must carry these to designated goal cells, then return the shelves to their original or empty locations. A new shelf is marked for delivery each time one is dropped off. This continues until the episode ends after $T$ timesteps. The reward is given by the total number of shelves delivered. We set $T=100$ and $n=2$ in all experiments, and vary $\ell$ and $m$ to modulate the difficulty of the tasks. We train the agents in the \emph{Small} setting, with $\ell=2$ and $m=32$.  

Formally, we define the domain as a MAPOMDP.
The state space $S$ includes all agents' coordinates and orientation, the positions of all shelves, the ID of all good shelves, and goal positions. 
The environment is partially observable, with each agent being able to observe only $d$ cells in every direction from their position. In all our experiments, we set $d=3$, making every observation a $7\times7$ box centered on the agent's position. 
The action space $A$ of each agent comprises four discrete actions: move forward, turn left/right, and load (which has the opposite effect if the agent is already carrying a shelf). 
The reward function $R$ is sparse: only when a good shelf is carried to a goal cell, the agents get a reward of $1$. 
This, together with the necessity to coordinate multiple agents in a non-stationary environment, makes the domain particularly challenging for RL.

According to the observations of the agents, we define the feature map $F_\mathcal{F}$ over the following predicates: \stt{ego(Status)}, \stt{good\_shelf(Dir,Dist)}, \stt{agent(Dir,Dist)}, \stt{goal(Dir,Dist)}, denoting, respectively: the status of the ego agent\footnote{The agent with respect to whom the observation was collected.}  (\stt{Status} $\in$ \stt{\{n,g,b\}}, meaning, respectively, that the ego agent is carrying no shelf, a good shelf or a bad shelf); the presence of a good shelf, or an agent, or a goal cell, in direction \stt{Dist} and at distance \stt{Dist}.
We set \stt{Dir} $\in$ \stt{\{north, south, east, west\}} and \stt{Dist} $\in \{0,\dotsc,3 \}$.
For this environment, we simply define $F_\mathcal{A}$ to map each action $a \in A$ to its equivalent in $\mathcal{H(A)}$ (i.e. $F_\mathcal{A}$ is bijective).

Given this formalization, we train MAPPO agents in the \emph{Small} configuration, consisting of an $11\times10$ grid featuring $32$ shelves, with $2$ of which being concurrently good, and extract logical policy representations on various checkpoints during training. We then compute the proposed metrics, repeating the procedure over 10 seeds.

We now report an example showing the theory learned by agent $1$ (best seed) at convergence ($3\cdot 10^7$ training steps, Figure \ref{fig:training_rware}):

\begin{verbatim}
forward :- V1 <= 1; ego(n); good_shelf(north, V1). 
forward :- V1 >= 3; ego(g); good_shelf(east, V1). 
forward :- V1 >= 0; ego(g); agent(east, V1). 
forward :- V1 >= 0; ego(b); agent(west, V1). 

left :- V2 <= 2; V1 <= 3; ego(n); good_shelf(south, V2); good_shelf(north, V1). 
left :- V2 <= 0; V1 >= 0; ego(n); good_shelf(south, V2); good_shelf(west, V1). 
left :- V1 <= 0; goal(south, V1); ego(b). 

right :- V1 >= 1; goal(east, V1); ego(g). 

load :- V2 <= 0; V1 <= 0; ego(n); good_shelf(east, V2); good_shelf(south, V1). 
load :- V2 <= 1; V1 >= 1; goal(west, V1); ego(b); agent(west, V2). 
\end{verbatim}

By inspecting this, we can see that we have learned human-readable and, apparently, meaningful rules. For example, for the action \stt{forward}, ILASP learns (first line) to take that action when the agent is not carrying anything (\stt{ego(n)}) and there is a good shelf (up to) distance 1 in the north direction. By taking \stt{forward}, the agent will actually move north one cell and will be able to load it. This is expressed by the second-to-last rule, which says that, when the agent is not carrying anything and is one cell directly below a good shelf, then it should load it. Also, the other rules seem of high quality: for example, the one for the action \stt{right}, which states that the agent should turn to the right when carrying a good shelf (\stt{ego(g)}) and a goal is within reach.

\subsection{Contrastive Multi-Agent Domain: Simple Adversary}\label{sec:mpe_formalization}
Simple Adversary \citep{lowe2017multi, mordatch2017emergence} (Figure \ref{fig:mpe}) is a contrastive, fully-observable multi-agent domain in which 2 good agents (light-blue circles) must prevent an adversary (red circle) from moving near the goal landmark (green dot), while trying, at the same time, to stay as close as possible to it. Only the good agents know which landmark is the goal one; the adversary does not. 

We model the domain as a MAMDP in which, for each agent, the state space $S$ consists of 5 $(x,y)$ tuples representing the \textit{relative} coordinates (w.r.t. the ego agent) of the goal landmark, of all landmarks, and of all the other agents.
Since the adversary does not know which landmark is the goal one, the coordinates of the goal landmark are set to zero in its own observation.
The action space $A$ comprises four discrete actions: move up, move down, move left, and move right.
The rewards are asymmetric and depend on the type of agents: good agents are positively rewarded based on their closest distance to the goal landmark, and penalized based on the adversary's distance to it. The adversary, instead, receives only a positive reward based on its distance to the target landmark. 

Features $\mathcal{F}$ include: \stt{agent\_0(Dir,Dist)}, \stt{agent\_1(Dir,Dist)}, \stt{landmark\_0(Dir,Dist)} and \stt{landmark\_1(Dir,Dist)}. These terms denote the direction and the relative distance, with respect to the agent for which they are defined, of the other agents and the two landmarks. For the `good' agents, the first agent term (\stt{agent\_0}) denotes the position and direction of the adversary, and the first landmark term (\stt{landmark\_0}) denotes the position and direction of the goal landmark.

We set \stt{Dir} $\in$ \stt{\{north, south, east, west\}} and \stt{Dist} $\in \{0,5,\dotsc,20 \}$. Since MDP state values (the coordinates of all agents and landmarks) are small floating numbers, we define $F_\mathcal{F}$ so that it scales them to obtain valid \stt{Dist} values by first multiplying each state by 10 and then truncating its values to the nearest integer.
Finally, in the same way as before, we define $F_\mathcal{A}$ as a straightforward translation of the MDP actions $a \in A$ to logical representations.

We train the MAPPO agents in the \emph{Normal} configuration, in which two landmarks (one of which is the goal) are assigned a random position in a $[-1,1] \times [-1,1]$ square.
The extracted metrics are then averaged over 5 seeds.

We now show an example of learned approximation of a `good agent' at convergence: 
%%% DO NOT REMOVE (agent_1, seed 102, 1000250)
\begin{verbatim}
move_down :- V1 <= 10; agent_0(south, V1). 

move_left :- V1 <= 15; landmark_0(west, V1). 

move_right :- V1 >= 5; landmark_0(east, V1). 
move_right :- V2 <= 0; V1 <= 0; agent_1(south, V2); landmark_1(east, V1). 
move_right :- V2 >= 0; V1 >= 10; V3 <= 5; agent_1(north, V3); landmark_1(south, V1); 
              landmark_0(east, V2). 
move_right :- V1 >= 0; V3 <= 10; V2 <= 15; agent_1(north, V3); landmark_1(west, V1); 
              landmark_0(east, V2). 

move_up :- V1 >= 5; landmark_1(north, V1). 

\end{verbatim}

The first line, for example, states that the agent should move down when \stt{agent\_0} (the adversary), is close in the south direction, to direct it away; or, it should move left (second line) when \stt{landmark\_0} (the target landmark) is west, etc. Generally, the learned approximations provide an interpretable representation of the most effective behaviour for the agent.

An example of a learned approximation of the `adversary' at convergence is the following:
\begin{verbatim}
move_down :- V2 >= 8; V1 >= 10; agent_0(east, V2); landmark_0(south, V1). 

move_left :- V1 >= 15; landmark_1(west, V1). 

move_right :- V1 <= 5; landmark_0(south, V1). 
move_right :- V1 <= 15; landmark_0(north, V1). 

move_up :- V2 >= 5; V1 >= 5; V3 <= 5; agent_1(south, V2); landmark_1(south, V3);
           landmark_0(south, V1). 
\end{verbatim}
In general, these rules direct the adversary to move towards a landmark, if an agent is not in the same direction (lines 1 and 2), and to avoid other agents, if they are in the way (last line, where both landmarks and an agent are south).

We now address each research question separately, presenting the most relevant results obtained with the proposed metrics. A complete set of metrics computed for each domain is reported in Appendix \ref{app:full_results}.

% ------------------ RQ1

\subsection{\RQinsighttraining, Explaining the training process}

\begin{figure}
    \centering
    % --- RWARE ---
    \begin{subfigure}[b]{0.49\linewidth}
        \centering
        \includegraphics[width=\linewidth]{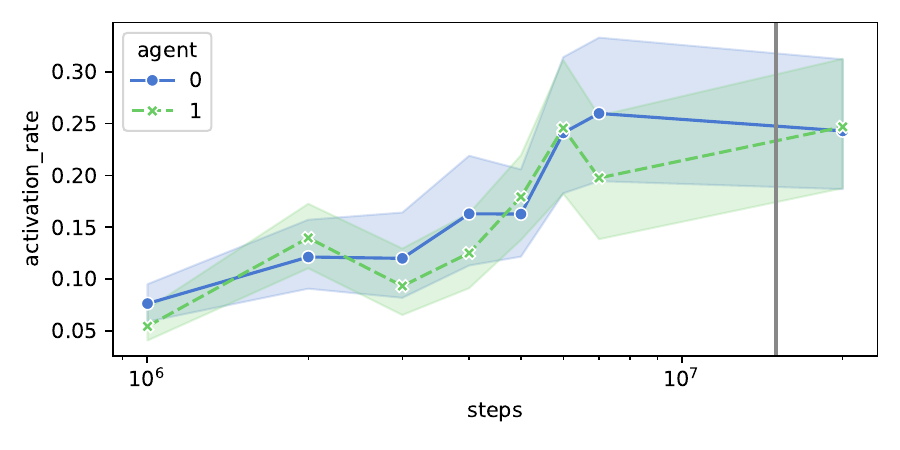}
        \caption{RWARE: activation rate for each agent.}  \label{fig:rware_activation_rate_agent}
    \end{subfigure}
    \hfill
    \begin{subfigure}[b]{0.49\linewidth}
        \centering
        \includegraphics[width=\linewidth]{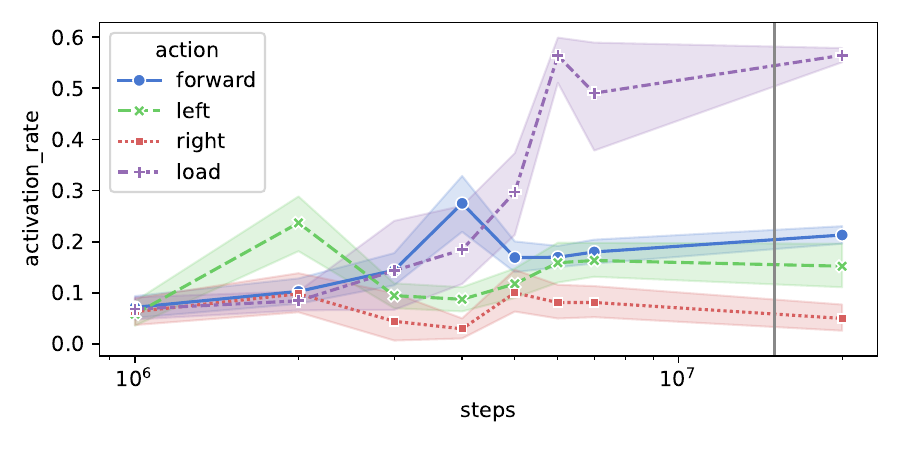}
        \caption{RWARE: activation rate averaged over agents.}
        \label{fig:rware_activation_rate_action}
    \end{subfigure}

    \vspace{1em}
    % --- Simple Adversary ---
    \begin{subfigure}[b]{0.49\linewidth}
        \centering
        \includegraphics[width=\linewidth]{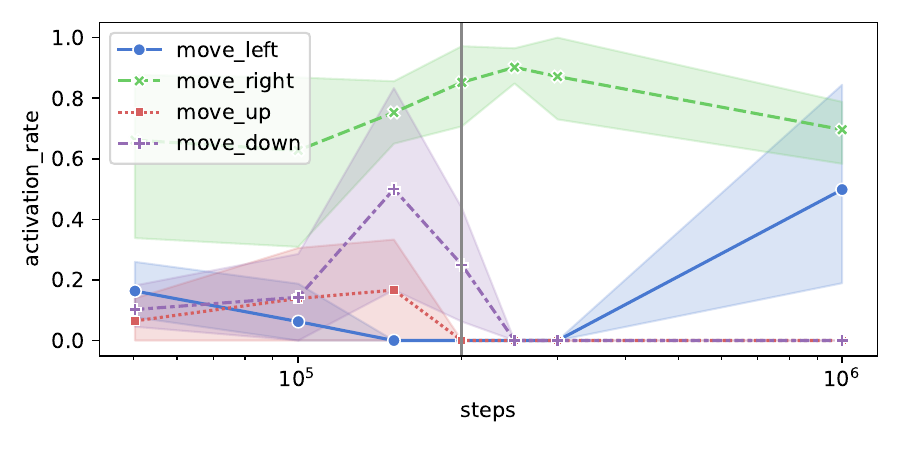}
        \caption{Simple Adversary: activation rate, adversary agent.}
        \label{fig:mpe_activation_rate_adv}
    \end{subfigure}
    \hfill
    \begin{subfigure}[b]{0.49\linewidth}
        \centering
        \includegraphics[width=\linewidth]{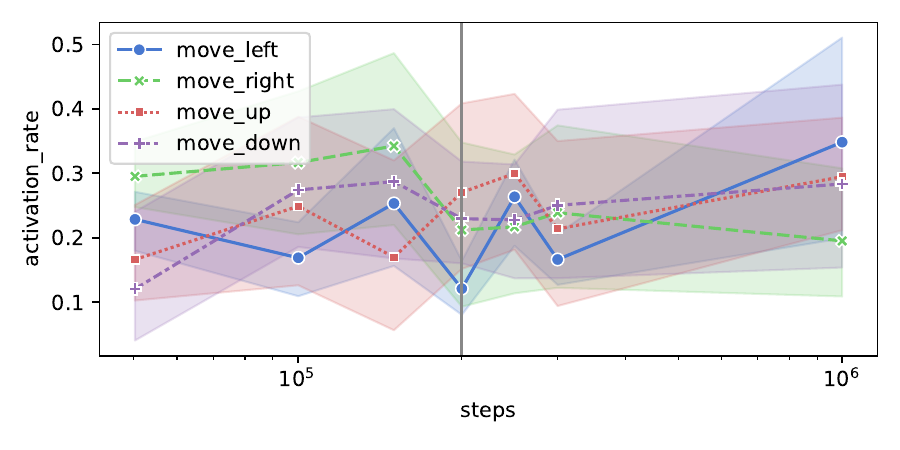}
        \caption{Simple Adversary: activation rate, good agents.}
        \label{fig:mpe_activation_rate_good}
    \end{subfigure}

    \caption{\textit{Activation rate} for RWARE (a–b) and Simple Adversary (c–d). Shaded areas represent the 95\% confidence interval over 10 theories for the RWARE domain and 5 theories for the Simple Adversary. The vertical gray line represents training curve convergence.}
    \label{fig:activation_rate}
\end{figure}

This research question aims to investigate how the proposed metrics can provide useful insights and additional information regarding the training process of RL agents, which we couldn't acquire by looking at the training curve only. To this aim, the activation rate (Figure \ref{fig:activation_rate}) is the most useful metric.

In Figure \ref{fig:rware_activation_rate_agent}, we can see the activation rate $\alpha(H)$ of the full logical theory for the two agents as the training progresses in the RWARE domain. Both increase consistently, reaching a peak of almost 30\% at convergence, from a low point of around 8\% at the start. This indicates that for both agents, as the training progresses, the symbolic representations align more and more with the behavior expressed by the policies, resulting in more accurate explanations at convergence (vertical line in the plot, see also Figure \ref{fig:training_rware}). Interestingly, the symbolic representation shows a convergence of the policy followed by the agents before the training curve converges (marked by the grey line).

Figure \ref{fig:rware_activation_rate_action} shows that we can gain even more useful information by looking at the activation rate for each action separately. In this case, for example, we notice that while the activation rate for the \stt{load} action grows significantly during training, the actions related to movements remain almost constant during training. This behavior can be explained by the fact that the \stt{load} action is much more specialized than movement actions. While movements can often be substituted or combined in different ways (e.g., turning right by performing three consecutive \stt{left} actions), \stt{load} has a unique and unambiguous outcome, and also plays a key role in maximizing and stabilizing the return (Figure \ref{fig:training_rware}). As a consequence, movement actions are harder to capture and describe within the logical representation, making their reliable learning more challenging. In this sense, the activation rates not only provide an estimate of how trustworthy the logical representations for each action are, but also offer an indication of how specialized and impactful each action is.
This interpretation is further supported by the results obtained in the Simple Adversary domain, where the \stt{left} and \stt{right} actions move the agent directly in the corresponding direction instead of rotating it. Moreover, in this setting, each movement becomes more critical for task completion, and indeed, the activation rate for these actions is significantly higher.
In more detail, in Figure \ref{fig:mpe_activation_rate_adv} we can see the activation rate $\alpha(\mathscr{a})$ for the first agent (the adversary). Interestingly, the activation rate for the actions \stt{move up} and \stt{move down} goes to zero after a certain point of the training. The reason is to be found in the state distribution. We find that, given the fact that the landmarks are often (almost) aligned on the horizontal axis (see Figure \ref{fig:mpe}, after a certain point, the adversary rarely moves up or down, only laterally. This shows that the good agents learn rapidly to cover well all the landmarks, leading the adversary to try going from one to another without success. Apparently, moving also vertically is not a good strategy for the adversary. Other than that, there is a clear bias toward moving right at the start for the adversary, which wanes at the end when $\alpha(\mathscr{a})$ for the two left and right actions approaches a very similar value. 
On the contrary, we can notice from Figure \ref{fig:mpe_activation_rate_good} that the activation rate for all the actions increases similarly, and there are no particular outliers like in the RWARE domain. In general, this indicates that all actions' rules are similarly coherent with the underlying policies of the good agents, probably due to well-balanced pick rates and informative traces, where no action is over-represented.
In general, like in RWARE, this metric provides insights about the training progress of the agents, without relying exclusively on the return, and identifying how the policies evolve even after the training curve reaches convergence.

\textbf{Overall, the activation rate proves to be a powerful diagnostic tool: it highlights when and how symbolic representations capture agent behavior, uncovers differences across the training process of specific actions (instead of the full policy measured by the return), and ultimately offers a richer perspective on the learning dynamics, beyond the apparent convergence of the return curve.}

% ------------------- RQ2

\subsection{\RQinsightmultiagent, Explaining multi-agent dynamics}

\begin{figure}
    \centering
    \begin{subfigure}[b]{0.49\linewidth}
        \centering
        \includegraphics[width=\linewidth]{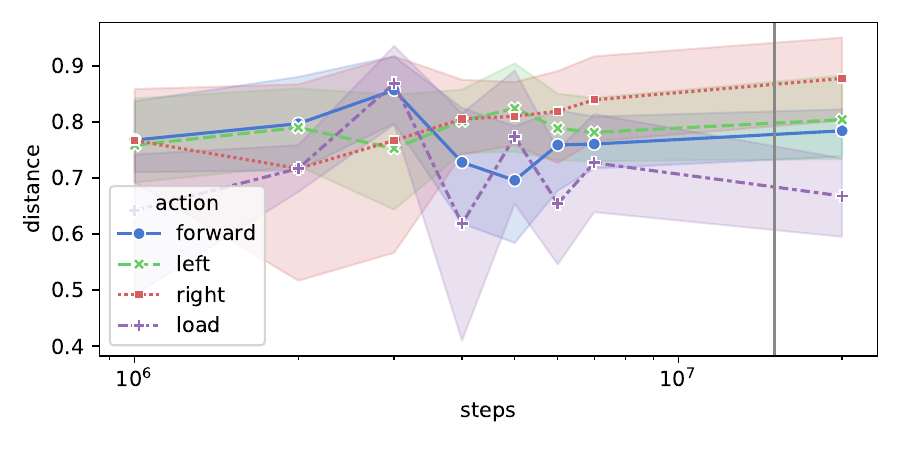}
        \caption{RWARE, syntactic distance for each action between theories of different agents.}
        \label{fig:syntactic_distance_agents}
    \end{subfigure}
    \hfill
    \begin{subfigure}[b]{0.49\linewidth}
        \centering
        \includegraphics[width=\linewidth]{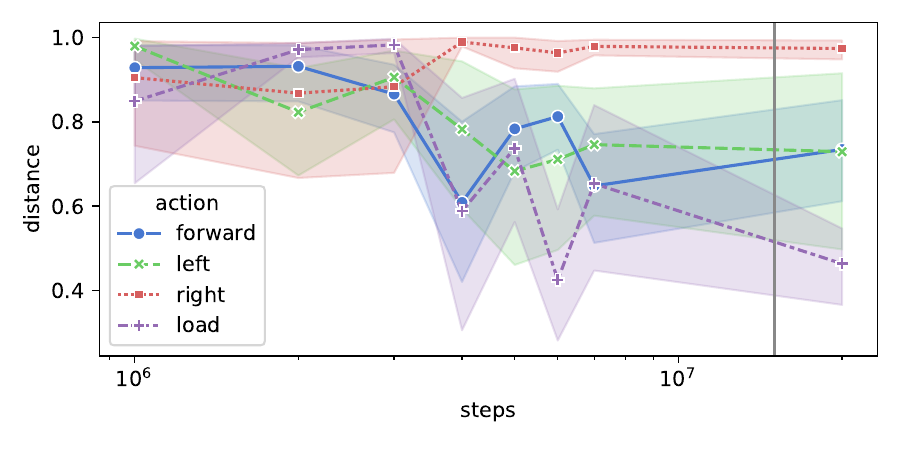}
        \caption{RWARE, semantic distance for each action between theories of different agents.}
        \label{fig:semantic_distance_actions}
    \end{subfigure}
    
    \begin{subfigure}[t]{0.49\linewidth}
        \centering
        \includegraphics[width=\linewidth]{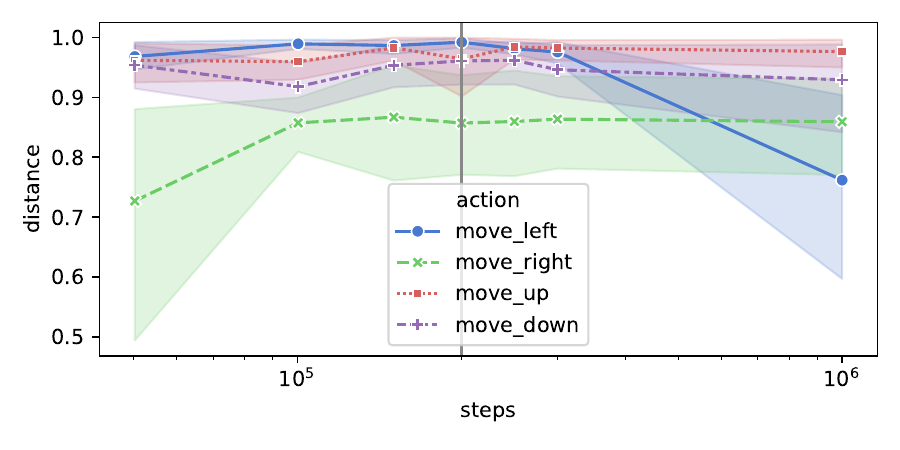}
        \caption{Simple Adversary, semantic distance between agents of different types.}\label{fig:mpe_semantic_dist_diff}
    \end{subfigure}
        \hfill
    \begin{subfigure}[t]{0.49\linewidth}
        \centering
        \includegraphics[width=\linewidth]{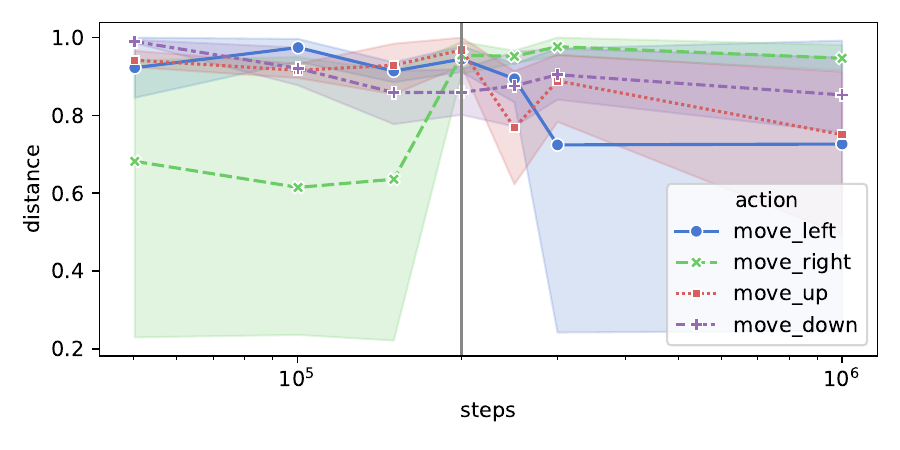}
        \caption{Simple Adversary, semantic distance between agents of the same type.}\label{fig:mpe_semantic_dist_same}
    \end{subfigure}
    
    \caption{Syntactic and semantic distances for RWARE and Simple Adversary under different conditions. Shaded areas represent the 95\% confidence interval. The vertical gray line represents training curve convergence.}
    \label{fig:distances}
\end{figure}

% A vital question in multi-agent scenarios is how to compare policies from different agents involved in the environment.
% While the activation rate of agents trained for the same tasks shows a similar trend (see Figure \ref{fig:activation_rate}), evidencing a potential agreement between them, we can gain more useful information while looking at the \textit{syntactic distance} and \textit{semantic distance} between them (Figure \ref{fig:distances}). 

A vital question in multi-agent scenarios is how to compare policies from different agents involved in the environment.
As we saw in Figure \ref{fig:activation_rate}, the activation rate of agents trained for the same tasks shows a broadly similar trend, suggesting some level of agreement. However, this indicator also highlights important differences: for instance, in the Simple adversary setting, comparing the activation rates of the adversary and the good agents reveals that certain actions carry very different importance for the two types of agents.
To this aim, to obtain a deeper understanding beyond this aggregate view, we can analyze the \textit{syntactic distance} and \textit{semantic distance} between policies (Figure \ref{fig:distances}), which provide complementary insights into how agents’ behaviors diverge or align.

Figure \ref{fig:syntactic_distance_agents} shows the average syntactic distance between the policies of the two RWARE agents during training. We observe that the distance does not significantly decrease over time. This effect is likely due to the growing divergence in the grounded terms used, which makes the two theories appear more dissimilar than they actually are.
In contrast, Figure \ref{fig:semantic_distance_actions} reports the semantic distance between the same two agents, computed separately for each action at different stages of training. As expected for agents trained to collaborate on the same task, the distance exhibits a clear decreasing trend, indicating that the two agents’ policies become increasingly aligned. The only exception is the \stt{right} action, which had already proved less reliable because of its low activation rate.

Focusing on the semantic distance, which is clearly more expressive and significant than the syntactic distance, a more interesting analysis can be performed on the Simple Adversary domain. Since agents are trained to pursue different objectives based on their types, we expect them to learn different behaviors. 

As expected, Figure \ref{fig:mpe_semantic_dist_diff} shows that the semantic distance between good and adversarial agents is very high, since in the same context they are required to adopt different strategies in order to achieve their respective goals. Interestingly, however, Figure \ref{fig:mpe_semantic_dist_same} clearly shows that, even though the good agents are trained for the same task, the resulting policies, and consequently their logical approximations, are semantically different, nearly as much as they are between adversarial agents. This is most likely due to the fact that the two agents develop complementary strategies to achieve the goal more effectively (e.g., each covering a different landmark), which makes their policies diverge despite being oriented toward the same objective.

Another important observation is that distances keep evolving even after the convergence of the training curve (grey line in the plots), signaling that, although the return stabilizes, the underlying policies continue to change and refine.

\textbf{Taken together, these results show that syntactic and semantic distances offer a unique perspective on inter-agent comparison in MARL, uncovering patterns of coordination, specialization, and adaptation of specific actions that remain invisible when looking only at the global return curve.}

% ------------------------- RQ3

\subsection{\RQfeatures, Explanations for evaluating domain features}

With this research question, we aim to go beyond the information that can be obtained through standard feature-importance methods for neural architectures, which are limited by treating the entire policy network as a single unit \cite{retzlaff2024human}. By contrast, the feature coverage metric that we introduce allows us to analyze the relevance of domain features on an action-by-action basis, thereby providing more fine-grained and detailed explanations.

\begin{figure}
    \centering
    \begin{subfigure}[b]{0.46\linewidth}
        \centering
        \includegraphics[width=\linewidth]{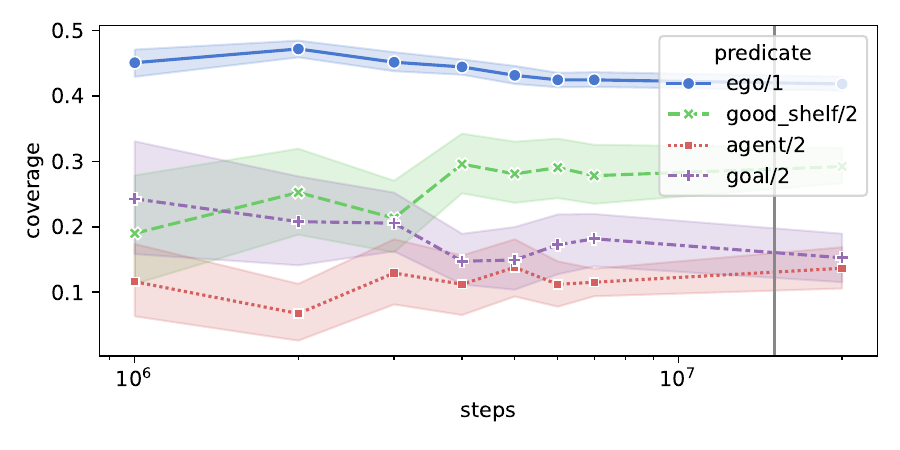}
        \caption{RWARE, feature coverage averaged over agent and over actions.}
        \label{fig:rware_feature_coverage}
    \end{subfigure}
    \hfill
    \begin{subfigure}[b]{0.49\linewidth}
        \centering
        \includegraphics[width=\linewidth]{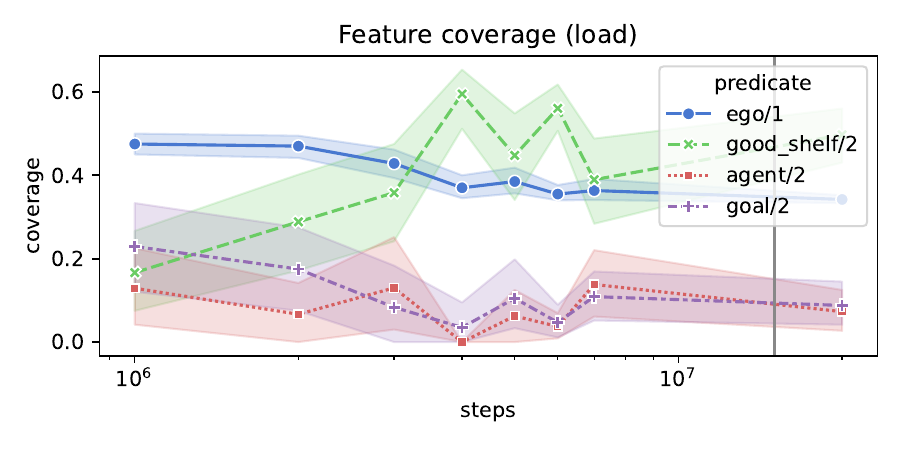}
        \caption{RWARE, feature coverage averaged over agents, \stt{load} action.}
        \label{fig:rware_feature_coverage_load}
    \end{subfigure}
    \begin{subfigure}[b]{0.49\linewidth}
        \centering
        \includegraphics[width=\linewidth]{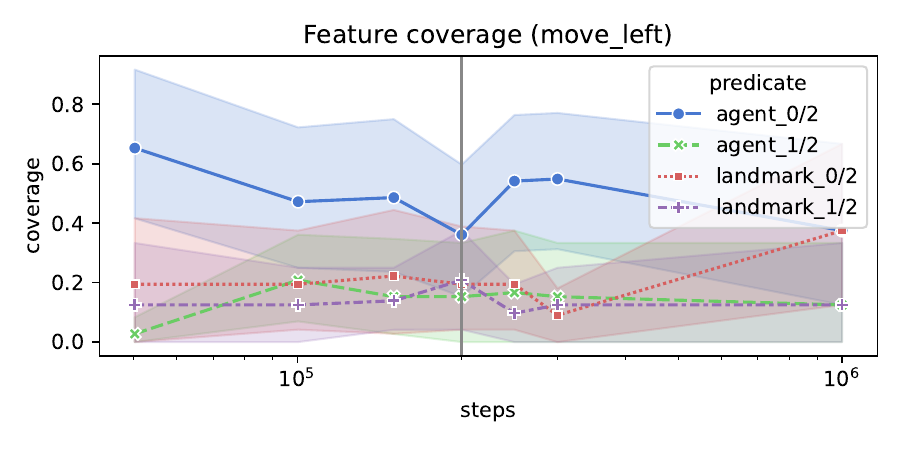}
        \caption{Simple Adversary, feature coverage averaged over agents, \stt{move\_left} action.}
        \label{fig:mpe_feature_coverage_left}
    \end{subfigure}
    \hfill
    \begin{subfigure}[b]{0.49\linewidth}
        \centering
        \includegraphics[width=\linewidth]{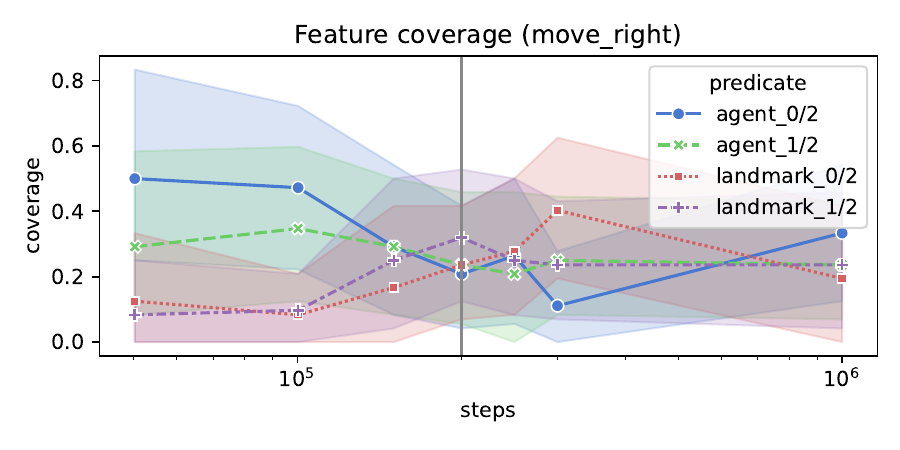}
        \caption{Simple Adversary, feature coverage averaged over agents, \stt{move\_right} action.}
        \label{fig:mpe_feature_coverage_right}
    \end{subfigure}
    
    \caption{Feature coverage averaged over all agents for the RWARE and Simple Adversary domains. Shaded areas represent the 95\% confidence intervals. The vertical gray line represents training curve convergence.}
    \label{fig:feature_coverage}
\end{figure}

Figure \ref{fig:rware_feature_coverage} shows the evolution of the feature coverage metric (averaged across agents) as a function of the training steps in the RWARE domain. Apart from the \stt{ego} predicate, which remains consistently impactful throughout training, the most represented predicates are \stt{good\_shelf}, whose importance increases with training, and \stt{goal}, which slightly decreases (from 25\% to 20\%) but nevertheless remains consistently relevant. This indicates that, as training progresses, agents become increasingly proficient at identifying good shelves, moving toward them, and picking them up.

An important advantage of feature coverage is that it allows us to zoom in on individual actions to discover specific behaviors. For instance, in the case of the \stt{load} action (Figure \ref{fig:rware_feature_coverage_load}), the predicate \stt{good\_shelf} emerges as particularly critical, much more than what is suggested by the feature importance averaged over the entire policy. 

Conversely, the predicate \stt{agent} remains relatively stable and overall less influential, since knowing the positions of other agents is typically less critical than information about shelves, carried items, or proximity to the goal, given that the reward is determined by the number of delivered good shelves.

In the Simple Adversary domain, however, the situation is different. We notice that, in this domain, all features have comparable importance, with the notable exception of \stt{agent\_0} (see Figure \ref{fig:mpe_feature_coverage_left} and Figure \ref{fig:mpe_feature_coverage_right} as an example for actions \stt{move\_left} and \stt{move\_down}). 
In fact, regardless of whether the metric is computed on a good or an adversarial agent (the figure reports the average across all agents), the predicate \stt{agent\_0} consistently refers to the agent of opposite type with respect to the one under analysis. It is therefore the most relevant feature to consider when deciding how to protect or reach the goal.

\begin{figure}
    \centering
    \begin{subfigure}[t]{0.49\linewidth}
        \centering
        \includegraphics[width=\linewidth]{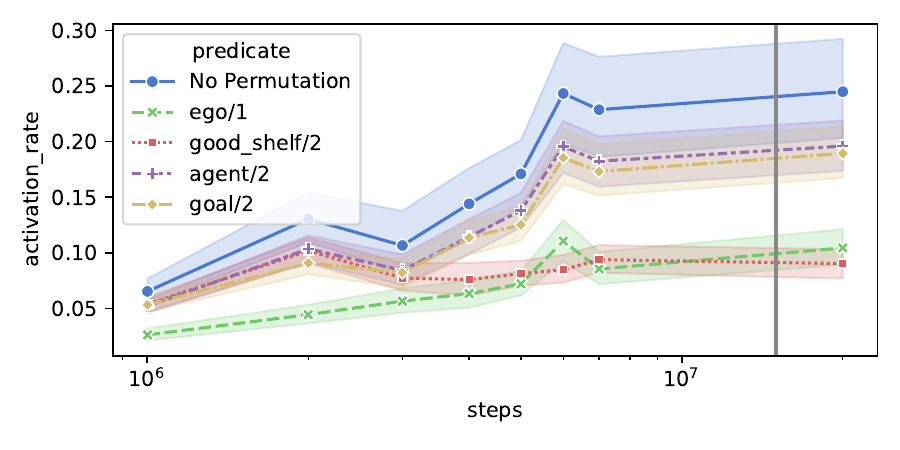}
        \caption{RWARE}\label{fig:rware_activation_perm}
    \end{subfigure}
    \hfill
    \begin{subfigure}[t]{0.49\linewidth}
        \centering
        \includegraphics[width=\linewidth]{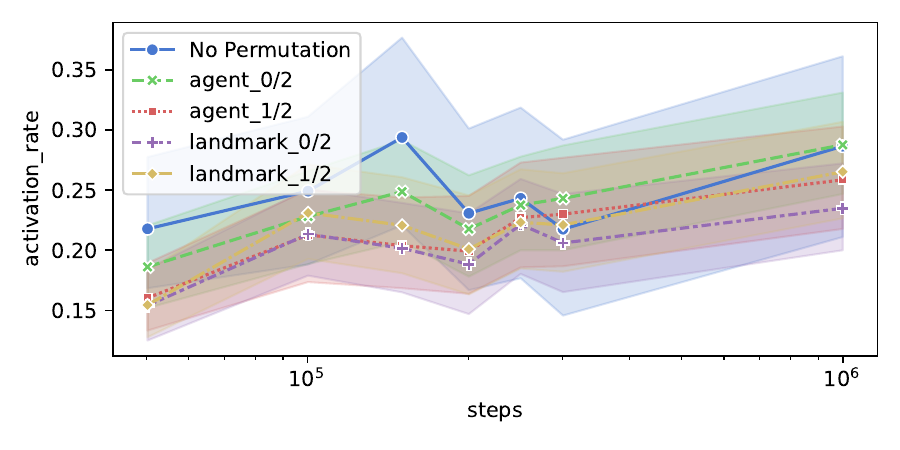}
        \caption{Simple Adversary}\label{fig:mpe_activation_perm}
    \end{subfigure}
    \caption{Activation rate (averaged over agents' theories) under \textit{constant permutation}, shaded areas represent 95\% confidence interval. The vertical gray line represents training curve convergence.}
    
    \label{fig:activation_perm}
\end{figure}

To evaluate each feature's importance in representing the policies, we also perform a \textit{constant permutation} procedure. That is, we randomly change the constants in ground feature terms (e.g., \stt{ego(g)} to \stt{ego(b)}) and recompute the activation rate after these changes. This allows us to test the policy approximations in terms of generalization and adherence to the original RL policy. 
Figure \ref{fig:rware_activation_perm} shows the activation rate for RWARE, after having applied the constant permutation procedure. The blue line is $\alpha(H)$ (averaged across the two agents' theories) before applying the permutation (hence over the original symbolic policies). We compare this with $\alpha(H)$ when permuting each single predicate (the other four lines). As expected, the activation rate of the original learned symbolic policy is generally higher. However, some predicates are more important than others, namely, \stt{ego} and \stt{good\_shelf}. Indeed, when permuted, they lead to a much greater deterioration of the metric. 

This suggests us that, if willing to decrease the number of features used to describe the domain, the \stt{agent} predicate could most likely be removed from $\mathcal{F}$, since it is the least influential in approximating the RL policies.

Figure \ref{fig:mpe_activation_perm}, on the other hand, shows that, in the Simple Adversary domain, feature permutation has a slighter impact on the activation rate of the whole theory, again confirming that all features have equal importance in describing the domain.

\textbf{Overall, these results show that feature coverage provides more fine-grained insights than standard feature-importance methods, since it allows us to disentangle the contribution of each domain feature not only globally but also at the level of individual actions. Moreover, the constant permutation analysis highlights which predicates are truly critical for policy approximation, offering a principled way to reduce the feature space without compromising explanatory power.}

% -------------- RQ4

\subsection{\RQknowledge, Explanations for transfer and generalization}
To evaluate the generalization and transferability of the learned theories beyond the training environment, we design experiments where symbolic policy representations learned in the reference settings (reported in Section \ref{sec:rware_formalization} for RWARE and Section \ref{sec:mpe_formalization} for Simple Adversary) are evaluated in different variants of the same domain. In practice, this is done by extracting the logical approximations at convergence in the baseline environments and computing the activation rate over execution traces collected in other environments with different levels of difficulty (Figure \ref{fig:activation_perm}).

In the case of RWARE, we consider four versions: \textit{Small} (with total number of shelves $m=32$, environment size $11 \times 10$, and number of concurrently good shelves $\ell=2$), \textit{Small/Hard} (same number of shelves and environment size, but $\ell = 1$), \textit{Large} ($m=80$, environment size $20 \times 10$, $\ell=2$) and \textit{Large/Hard} (same number of shelves and environment size as the large version, but $\ell = 1$).

\begin{figure}
    \centering
    \begin{subfigure}[t]{0.49\linewidth}
        \centering
        \includegraphics[width=\linewidth]{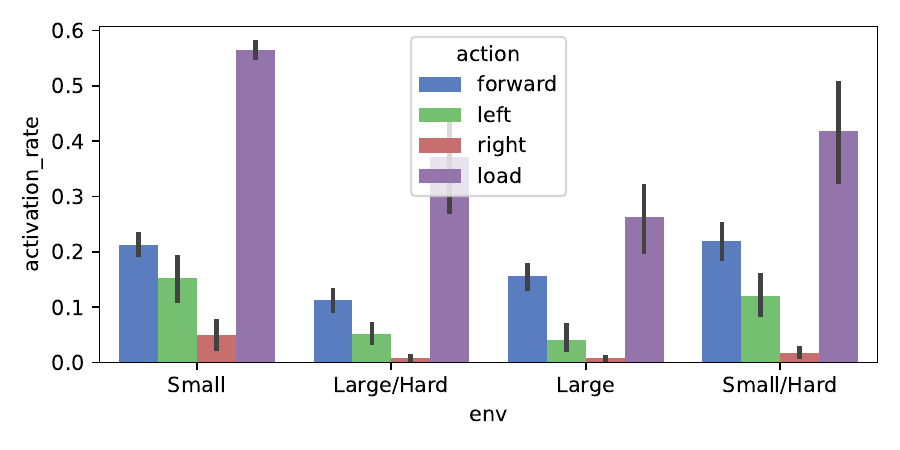}
        \caption{RWARE}\label{fig:rware_activation_gen}
    \end{subfigure} 
    \begin{subfigure}[t]{0.49\linewidth}
        \centering
        \includegraphics[width=\linewidth]{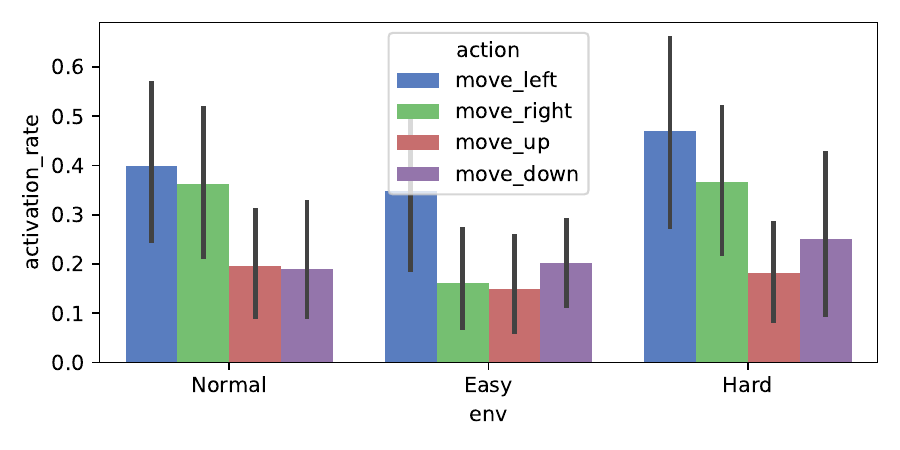}
        \caption{Simple Adversary}\label{fig:mpe_activation_gen}
    \end{subfigure}

    \caption{Generalization of activation rate of the learned logical policy representations at convergence using different environment versions to collect execution traces and contexts. The activation rate is averaged over agents.}
    \label{fig:rware_generalization}
\end{figure}

Figure \ref{fig:rware_activation_gen} shows the activation rate when different environments are used to generate the contexts rather than the one used to learn the policy approximations. That is, we use the symbolic rules learned at convergence in the \textit{Small} environment (see Section \ref{sec:rware_formalization}) but test them by collecting execution traces in different (more challenging) RWARE environments (\textit{Hard}, \textit{Large}, \textit{Large/Hard}). For reference, we also include the results for the baseline environment (\textit{Small}). Generally, we can see that the activation rate for each action decreases when testing other environments, particularly when increasing the grid size (\textit{Large} variants). However, we can still see how the activation rate for the action \stt{load} remains pretty high in every test, confirming that the rules regarding this action are the most reliable and thus represent potential candidates for transferring useful symbolic knowledge outside of the training setting, to help scale to more challenging scenarios.

For the Simple Adversary domain, we test the generalization of the learned policy approximations across different settings. Besides the \textit{Normal} version (see Section \ref{sec:mpe_formalization}), where at the beginning of each episode the two landmarks are randomly placed within the square $[-1,1] \times [-1,1]$, we also consider an \textit{Easier} variant (landmarks sampled in the smaller square $[-0.5,0.5] \times [-0.5,0.5]$) and a \textit{Harder} one (landmarks sampled in the larger square $[-2,2] \times [-2,2]$). These variants differ in difficulty, especially for the good agents: when landmarks are closer together, they are easier to cover, whereas larger distances make them harder to control and easier for the opponent to disrupt. This intuition is confirmed by the training curves in Figure \ref{fig:training_mpe}.

In this case, as shown in Figure \ref{fig:mpe_activation_gen}, we see less difference between computing the activation rate on different environments than was the case for RWARE (compare Figure \ref{fig:rware_activation_gen}). This suggests that the learned theories at convergence are quite generalizable and can be used to extract useful knowledge that could be exploited in similar domains.

\textbf{Overall, these experiments highlight how the activation rate provides a practical and interpretable metric to assess not only the reliability of the learned rules within the training domain but also their potential for transferability across different settings.}

\section{Conclusions and Future Works}\label{sec:conclusions}

In this work, we advance the state of the art in explainable RL and logics for explainable AI. Specifically, we introduce a novel set of objective, user-independent explainability metrics for RL agents, specifically designed to evaluate symbolic policy representations. 
%By leveraging ILP to extract logical policies from agent behavior, we provide a framework to systematically assess different aspects of policy explainability, including confidence, feature relevance, and policy evolution beyond the measurement of the final return curves, thus gaining extra information about the training process. 
By leveraging ILP to extract logical policies from agent behavior, we provided a framework to systematically assess different aspects of policy explainability, including confidence, understood as the degree to which symbolic rules reliably align with the agent’s actual decisions and reflect stable behavioral patterns across different states, feature relevance, and policy evolution beyond the measurement of the final return curves, thus gaining extra information about the training process. 

Our empirical evaluation across single-agent and multi-agent RL scenarios demonstrates the practical utility of the proposed metrics. Specifically, the \emph{activation rate} proves to be a powerful diagnostic tool, revealing not only how well the symbolic rules capture the agent’s behavior but also highlighting action-specific learning dynamics that can't be obtained when evaluating only the global return. Similarly, \emph{feature coverage} provides fine-grained insights into the contribution of individual environmental features, enabling principled feature selection and enhancing interpretability at both the global and action-specific levels. Finally, \emph{syntactic} and \emph{semantic distances} offer a novel perspective for comparing policies, uncovering patterns of coordination, specialization, and adaptation in multi-agent settings that remain invisible when considering conventional performance metrics alone. Together, these metrics facilitate a more complete understanding of policy behavior, robustness, and generalization capabilities.

Beyond these findings, our results suggest that explainability metrics can be effectively integrated into the RL training and evaluation loop to support debugging, policy refinement, and transfer learning. By quantifying the alignment between symbolic rules and agent behavior, our framework provides a principled approach to assess the reliability and potential transferability of learned skills across different environments.

As future work, several directions can further enhance the applicability and scope of this work. First, extending the logical formalism to more refined representations, including temporal and sequential dependencies, could provide deeper insights into policy dynamics in long-horizon tasks. Second, exploring novel ways to leverage the symbolic knowledge acquired in RL may improve both policy optimization and transferability across environments. Third, integrating user-centric evaluations with our objective metrics could support adaptive explainability tools tailored to diverse stakeholders, including domain experts and end-users. Finally, applying the proposed framework to more complex, real-world multi-agent systems, including human-in-the-loop scenarios, would allow validating its effectiveness in safety-critical and more complex (e.g., robotics) domains.

In conclusion, our work provides a fundamental step toward rigorous, objective, and actionable explainability in RL, offering tools to better understand, trust, and refine autonomous agents.

%%
%% The acknowledgments section is defined using the "acks" environment
%% (and NOT an unnumbered section). This ensures the proper
%% identification of the section in the article metadata, and the
%% consistent spelling of the heading.

%%
%% The next two lines define the bibliography style to be used, and
%% the bibliography file.
\bibliographystyle{ACM-Reference-Format}
\bibliography{sample-base}

%%
%% If your work has an appendix, this is the place to put it.
\appendix

\section{Complete results for the experimental evaluation}\label{app:full_results}
The following presents the complete results of computing the proposed metrics in various settings. 
% ------------------- RWARE

\subsection{RWARE}

\begin{figure}[h]
    \centering
    \begin{subfigure}[b]{0.4\linewidth}
        \centering
        \includegraphics[width=\linewidth]{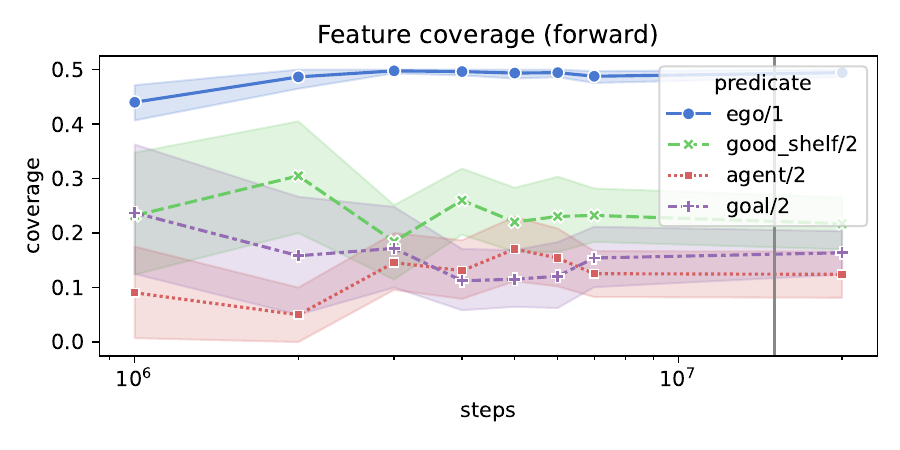}
        %\caption{\stt{forward}} 
    \end{subfigure}
    \hspace{.7cm}
    \begin{subfigure}[b]{0.4\linewidth}
        \centering
        \includegraphics[width=\linewidth]{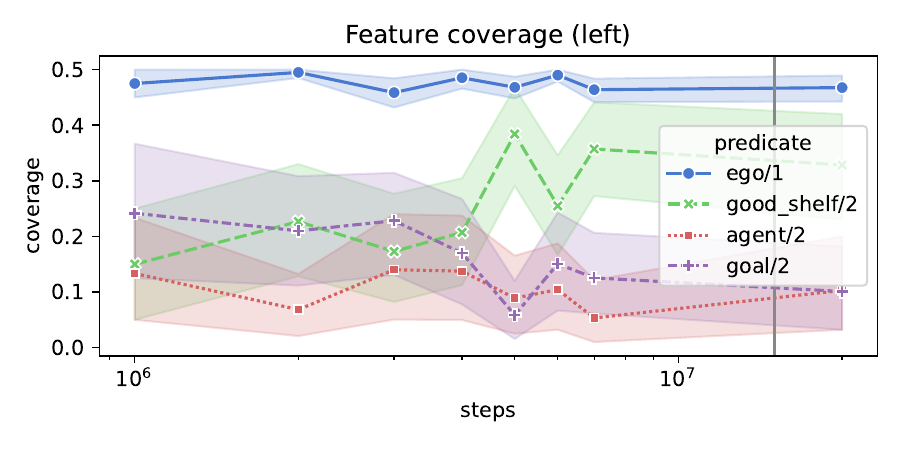}
        %\caption{\stt{left}}
    \end{subfigure}
    
    \begin{subfigure}[b]{0.4\linewidth}
        \centering
        \includegraphics[width=\linewidth]{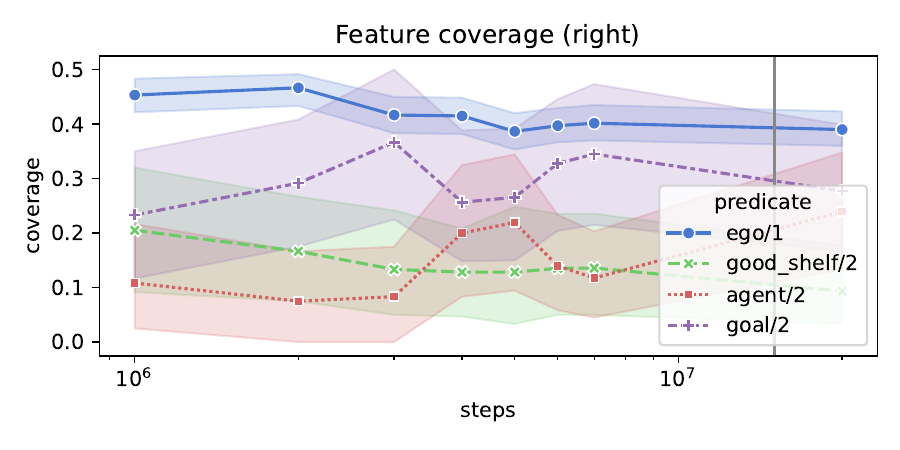}
        %\caption{\stt{right}}
    \end{subfigure}
    \hspace{.7cm}
    \begin{subfigure}[b]{0.4\linewidth}
        \centering
        \includegraphics[width=\linewidth]{images/rware/feature_coverage_action_load.pdf}
        %\caption{\stt{load}}
    \end{subfigure}

    \caption{Feature coverage for each action in the RWARE domain.}
    \label{fig:app_feature_coverage}
\end{figure}

\begin{figure}[h]
    \centering
    \begin{subfigure}[b]{0.4\linewidth}
        \centering
        \includegraphics[width=\linewidth]{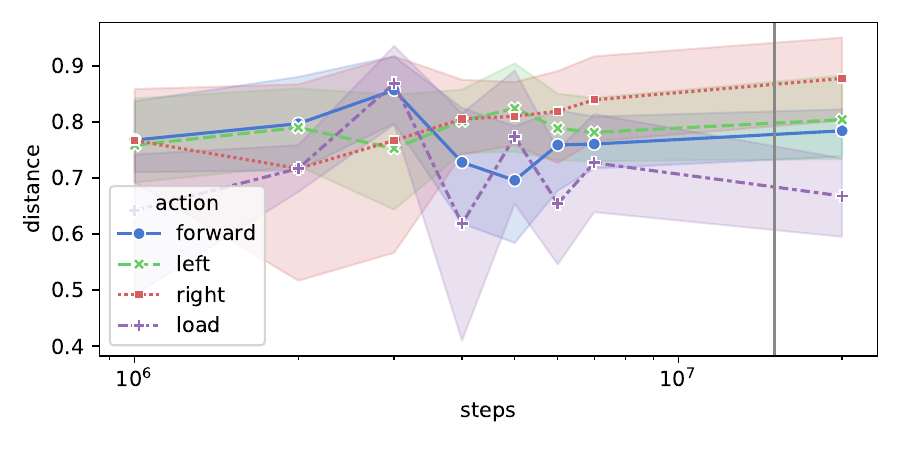}
        %\caption{Syntactic distance} 
    \end{subfigure}
    \hspace{.7cm}
    \begin{subfigure}[b]{0.4\linewidth}
        \centering
        \includegraphics[width=\linewidth]{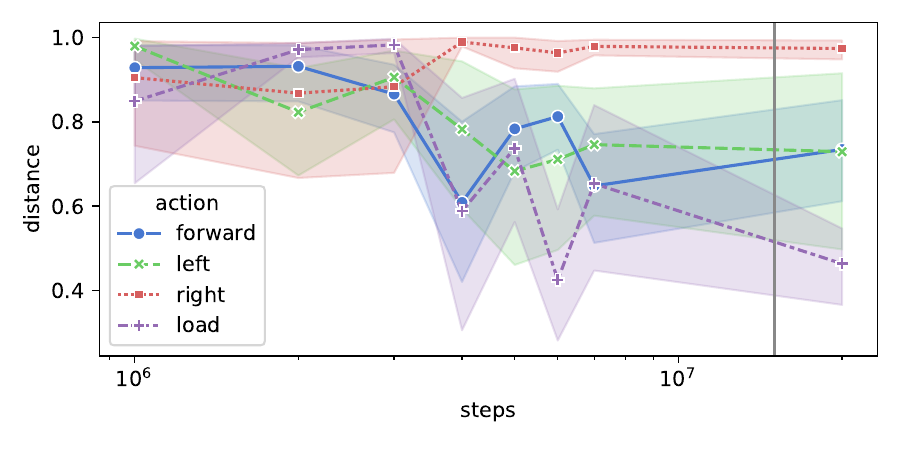}
        %\caption{Semantic distance}
    \end{subfigure}

    \caption{Syntactic (left) and semantic (right) distance between agents in the RWARE domain.}
    \label{fig:app_syn_sem_dist}
\end{figure}

\begin{figure}[h]
    \centering
    \begin{subfigure}[b]{0.4\linewidth}
        \centering
        \includegraphics[width=\linewidth]{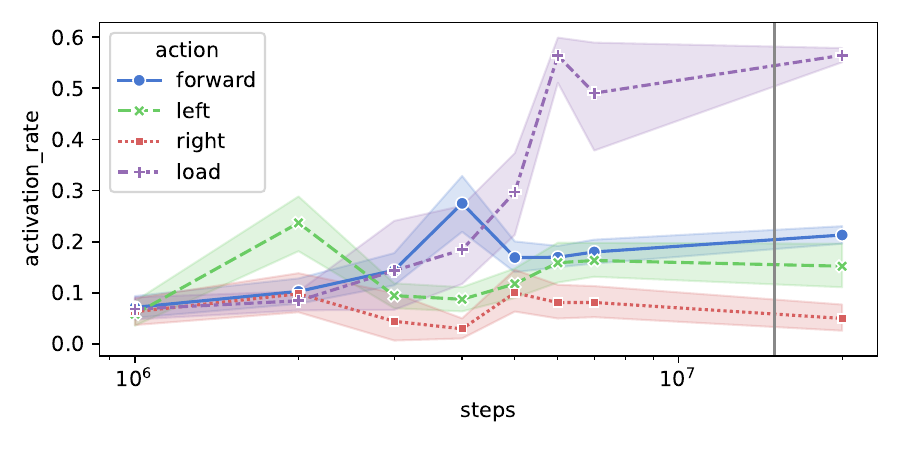}
        %\caption{Syntactic distance} 
    \end{subfigure}
    \hspace{.7cm}
    \begin{subfigure}[b]{0.4\linewidth}
        \centering
        \includegraphics[width=\linewidth]{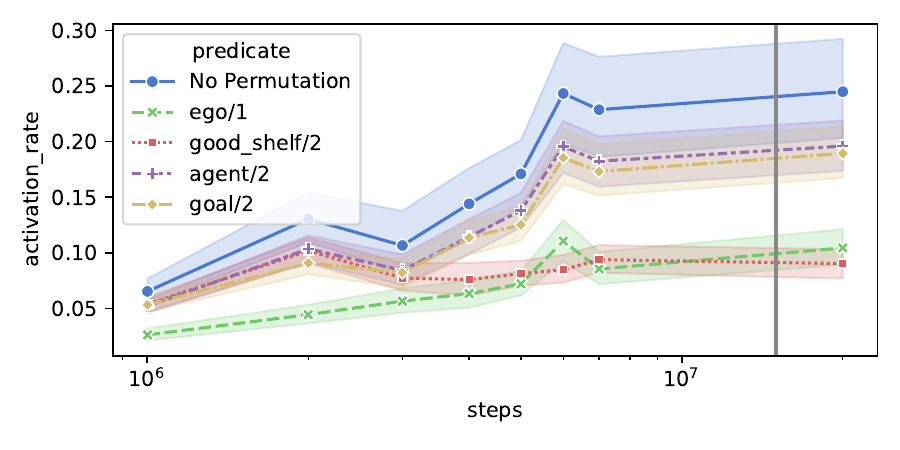}
        %\caption{Semantic distance}
    \end{subfigure}

    \caption{Activation rate, standard (left) and under constant permutation (right) in the RWARE domain.}
    \label{fig:app_activation_rate}
\end{figure}

% --------------------- Simple Adversary

\subsection{Simple Adversary}

\begin{figure}[h]
    \centering
    \begin{subfigure}[b]{0.4\linewidth}
        \centering
        \includegraphics[width=\linewidth]{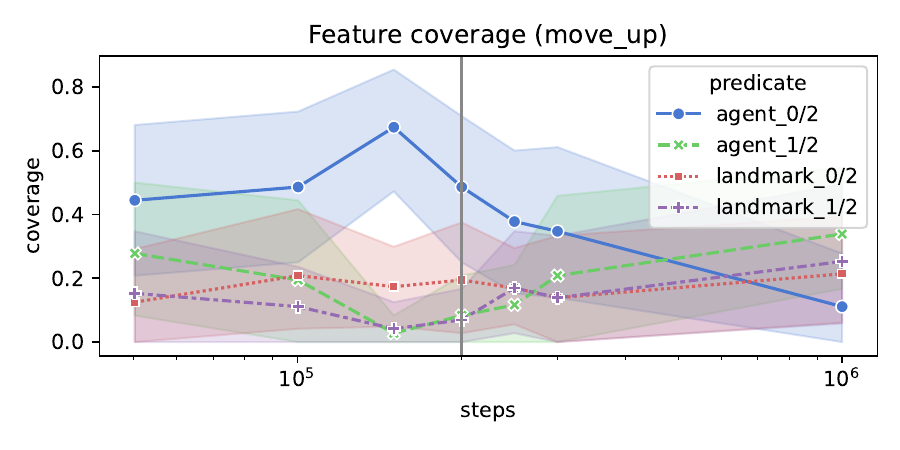}
        %\caption{\stt{forward}} 
    \end{subfigure}
    \hspace{.7cm}
    \begin{subfigure}[b]{0.4\linewidth}
        \centering
        \includegraphics[width=\linewidth]{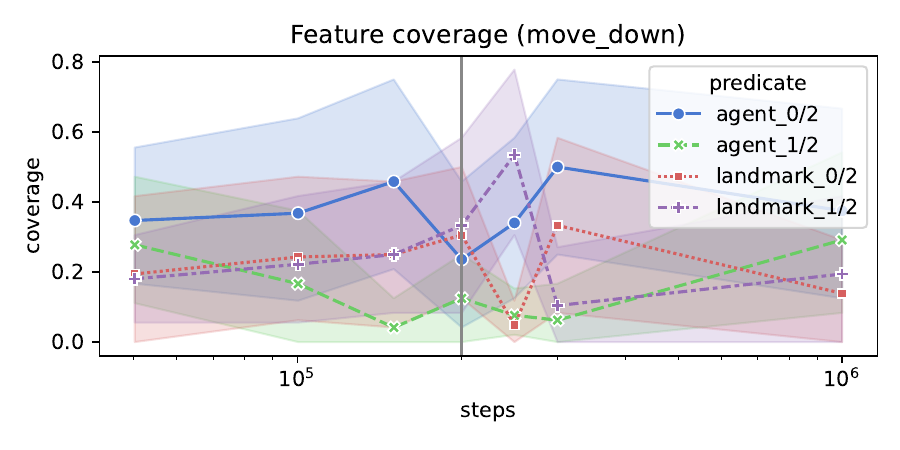}
        %\caption{\stt{left}}
    \end{subfigure}
    
    \begin{subfigure}[b]{0.4\linewidth}
        \centering
        \includegraphics[width=\linewidth]{images/MPE/feature_coverage_action_move_left.pdf}
        %\caption{\stt{right}}
    \end{subfigure}
    \hspace{.7cm}
    \begin{subfigure}[b]{0.4\linewidth}
        \centering
        \includegraphics[width=\linewidth]{images/MPE/feature_coverage_action_move_right.pdf}
        %\caption{\stt{load}}
    \end{subfigure}

    \caption{Feature coverage for each action in the Simple Adversary domain.}
    \label{fig:app_feature_coverage_adv}
\end{figure}

\begin{figure}[h]
    \centering
    \begin{subfigure}[b]{0.4\linewidth}
        \centering
        \includegraphics[width=\linewidth]{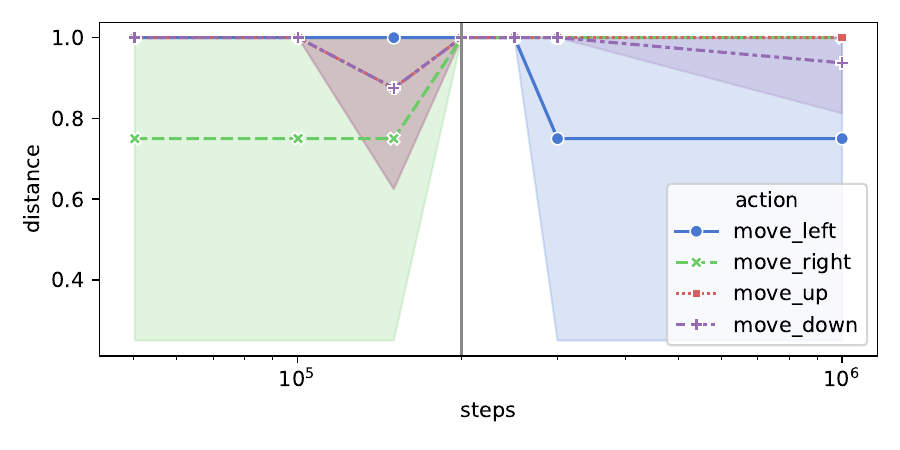}
        %\caption{Syntactic distance} 
    \end{subfigure}
    \hspace{.7cm}
    \begin{subfigure}[b]{0.4\linewidth}
        \centering
        \includegraphics[width=\linewidth]{images/MPE/semantic_distance_actions_same_agents.pdf}
        %\caption{Semantic distance}
    \end{subfigure}
    
    \begin{subfigure}[b]{0.4\linewidth}
        \centering
        \includegraphics[width=\linewidth]{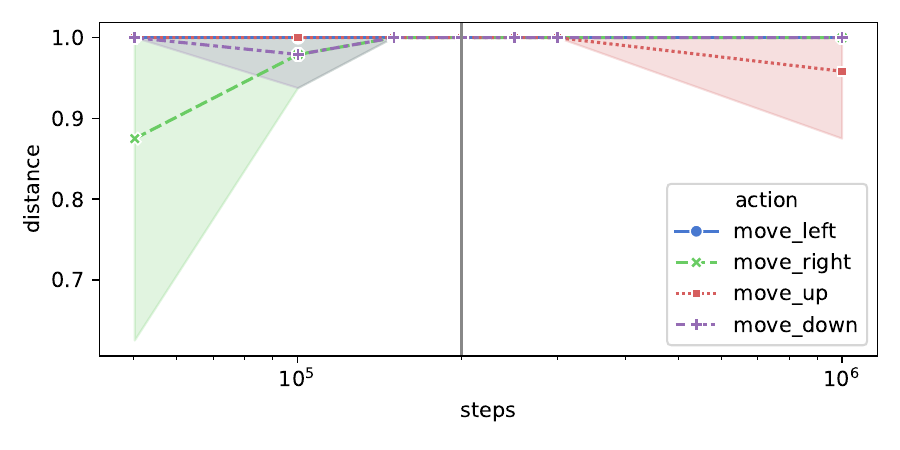}
        %\caption{Syntactic distance} 
    \end{subfigure}
    \hspace{.7cm}
    \begin{subfigure}[b]{0.4\linewidth}
        \centering
        \includegraphics[width=\linewidth]{images/MPE/semantic_distance_actions_diff_agents.pdf}
        %\caption{Semantic distance}
    \end{subfigure}

    \caption{Syntactic (left) and semantic (right) distance between agents of the same type (first row) and of different types (second row) in the Simple Adversary domain.}
    \label{fig:app_syn_sem_dist_adv}
\end{figure}

\begin{figure}[h]
    \centering
    \begin{subfigure}[b]{0.4\linewidth}
        \centering
        \includegraphics[width=\linewidth]{images/MPE/activation_rate_agent_0.pdf}
        %\caption{Syntactic distance} 
    \end{subfigure}
    \hspace{.7cm}
    \begin{subfigure}[b]{0.4\linewidth}
        \centering
        \includegraphics[width=\linewidth]{images/MPE/activation_rate_agent_same.pdf}
        %\caption{Semantic distance}
    \end{subfigure}

    \begin{subfigure}[b]{0.4\linewidth}
        \centering
        \includegraphics[width=\linewidth]{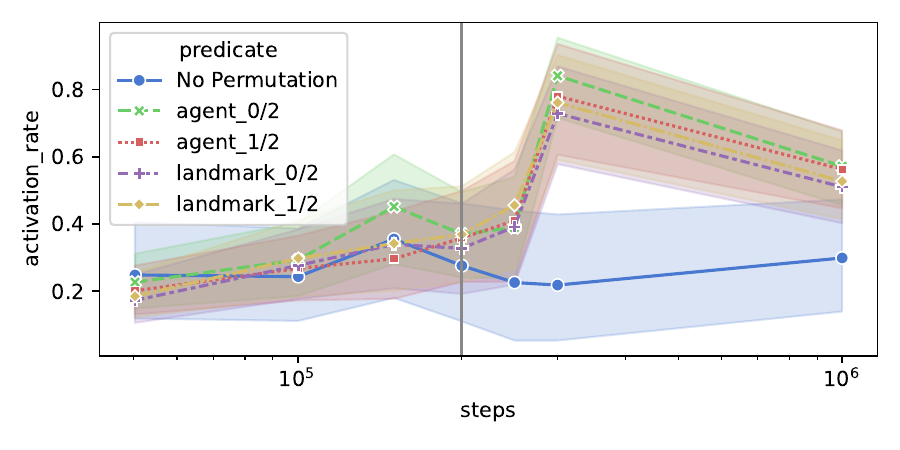}
        %\caption{Syntactic distance} 
    \end{subfigure}
    \hspace{.7cm}
    \begin{subfigure}[b]{0.4\linewidth}
        \centering
        \includegraphics[width=\linewidth]{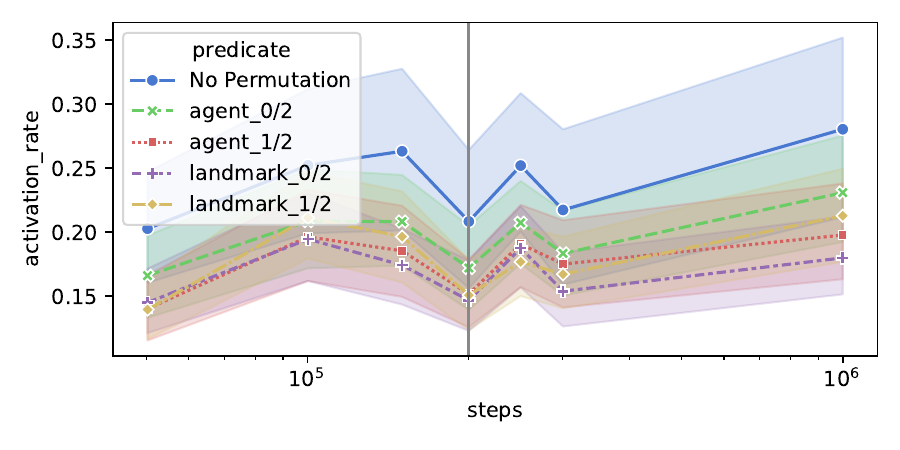}
        %\caption{Semantic distance}
    \end{subfigure}

    \caption{Activation rate of adversary (left) and good agents (right) in the Simple Adversary domain. Standard (first row) and under permutation (second row).}
    \label{fig:app_activation_rate_adv}
\end{figure}

\end{document}